\documentclass{article}

\usepackage{microtype}
\usepackage{graphicx}
\usepackage{subcaption}
\usepackage{booktabs} 

\usepackage{hyperref}

\usepackage[preprint]{icml2026}

\usepackage{amsmath,amsfonts,bm}

\def\eqref#1{equation~\ref{#1}}

\def\1{\bm{1}}

\usepackage{xcolor}

\newif\ifshowcomments

\DeclareMathAlphabet{\mathsfit}{\encodingdefault}{\sfdefault}{m}{sl}
\SetMathAlphabet{\mathsfit}{bold}{\encodingdefault}{\sfdefault}{bx}{n}

\usepackage{amsmath}
\usepackage{amssymb}
\usepackage{mathtools}
\usepackage{amsthm}
\usepackage{multirow}
\usepackage{hyperref}
\usepackage{url}
\usepackage{graphicx}
\usepackage{subcaption}
\usepackage{diagbox}
\usepackage{multirow}
\usepackage{makecell}
\usepackage{booktabs}
\usepackage{siunitx}
\usepackage{cancel}
\usepackage{cleveref}
\usepackage{float}
\usepackage{algorithm}
\usepackage[noend]{algpseudocode}
\algrenewcommand\alglinenumber[1]{\scriptsize #1:}
\usepackage[most]{tcolorbox}
\usepackage{xcolor}

\usepackage{tabularx} 
\usepackage{makecell} 
\usepackage{booktabs} 
\usepackage{siunitx}  
\definecolor{darkgreen}{rgb}{0.0, 0.5, 0.0}
\definecolor{lightblue}{rgb}{0.4, 0.7, 0.95}

\theoremstyle{plain}

\theoremstyle{definition}

\theoremstyle{remark}

\usepackage[disable,textsize=tiny]{todonotes}

\usepackage[table]{xcolor}
\definecolor{lightgreen}{RGB}{220,245,220} 
\newcommand{\bestcell}[1]{\cellcolor{lightgreen}{#1}}

\icmltitlerunning{ Learning Adaptive LLM Decoding }

\begin{document}
\twocolumn[
  \icmltitle{ Learning Adaptive LLM Decoding }



  \icmlsetsymbol{intern}{$\dagger$}

  \begin{icmlauthorlist}
    \icmlauthor{Chloe H. Su}{sch,sch1,intern}
    \icmlauthor{Zhe Ye}{sch2}
    \icmlauthor{Samuel Tenka}{comp}
    \icmlauthor{Aidan Yang}{comp}
    \icmlauthor{Soonho Kong}{comp}
    \icmlauthor{Udaya Ghai}{comp}
  \end{icmlauthorlist}

  \icmlaffiliation{sch}{Department of Computer Science, Harvard University, Boston, USA}
  \icmlaffiliation{comp}{Amazon, Boston, MA 02210, USA}
  \icmlaffiliation{sch1}{Kempner Institute, Harvard University, Cambridge, MA, USA}
  \icmlaffiliation{sch2}{UC Berkeley, CA, USA}

  \icmlcorrespondingauthor{Chloe Su}{csu@g.harvard.edu}
  \icmlcorrespondingauthor{Udaya Ghai}{ughai@amazon}
  \icmlkeywords{Machine Learning, ICML}

  \vskip 0.3in
]
\printAffiliationsAndNotice{\textsuperscript{$\dagger$}Work done during internship at Amazon.}


\begin{abstract}

Decoding from large language models (LLMs) typically relies on fixed sampling
hyperparameters (e.g., temperature, top-$p$), despite substantial variation in
task difficulty and uncertainty across prompts and individual decoding steps. We propose to learn
adaptive decoding policies that dynamically select sampling strategies at
inference time, conditioned on available compute resources. Rather than
fine-tuning the language model itself, we introduce lightweight decoding
adapters trained with reinforcement learning and verifiable terminal rewards
(e.g.\ correctness on math and coding tasks). At the sequence level, we frame
decoding as a contextual bandit problem: a policy selects a decoding strategy
(e.g.\ greedy, top-$k$, min-$p$) for each prompt, conditioned on the prompt
embedding and a parallel sampling budget. At the token level, we model decoding
as a partially observable Markov decision process
(POMDP), where a policy selects sampling actions at each token step based on
internal model features and the remaining token budget. Experiments on the MATH
and CodeContests benchmarks show that the learned adapters improve the
accuracy–budget tradeoff: on MATH, the token-level adapter improves Pass@1
accuracy by up to 10.2\% over the best static baseline under a fixed token
budget, while the sequence-level adapter yields 2–3\% gains under fixed parallel
sampling. Ablation analyses support the contribution of both sequence- and
token-level adaptation.

\end{abstract}
\section{Introduction}

Large language models (LLMs) have achieved remarkable performance across domains
ranging from mathematical reasoning~\cite{ren2025deepseek,
yang2024qwen25mathtechnicalreportmathematical} to code
generation~\cite{chen2021evaluating, princis2025treecoder} and scientific
discovery~\cite{boiko2023autonomous}. Despite this progress, inference from LLMs
remains a major computational bottleneck. A key source of inefficiency lies in
decoding, the process by which a model samples output tokens from its predictive
distribution. In current practice, decoding relies on fixed sampling
hyperparameters such as temperature, top-$k$, and top-$p$, chosen statically for
an entire model or dataset. This ignores the substantial heterogeneity across
prompts, reasoning styles, and even individual tokens. In many cases, the
optimal decoding strategy depends on latent features such as token-level
uncertainty or problem structure, with recent analyses showing that uncertainty
during reasoning is often concentrated at a small number of high-entropy
tokens~\cite{lin2024critical, wang2025beyond}.

Recent work has explored adaptive sampling and confidence-aware decoding,
demonstrating that modulating stochasticity can substantially affect generation
quality and reasoning performance~\cite{nguyen2024turning, zhang2024edt,
dhuliawala2024adaptive, lin2024critical, wang2025beyond}. However, these
approaches typically rely on static heuristics or offline-tuned parameters, and
do not incorporate decoding decisions directly into an end-to-end learning
objective.

In parallel, reinforcement learning with verifiable rewards (RLVR) has emerged
as a powerful framework for improving reasoning performance in large language
models~\cite{guo2025deepseek}. However, in many RLVR pipelines, decoding
strategy is treated as fixed during generation, even though the choice of
sampling method directly influences the support and diversity of generated
outputs. One contributing factor is that commonly used decoding strategies such
as top-$k$ or nucleus sampling modify the support of the output distribution,
which can complicate their integration into standard policy-gradient training.

As a result, in widely used open-source RLVR frameworks, decoding
hyperparameters such as temperature, top-$k$, and top-$p$ are typically treated
as fixed generation settings rather than learnable or adaptive components of the
policy~\cite{vonwerra2020trl}. In practice, these hyperparameters are often
adjusted post hoc at inference time to trade off accuracy, diversity, and
computational cost. This separation can induce a train--test mismatch, in which models are optimized under a fixed decoding distribution and budget, but evaluated or deployed under different inference-time constraints. In our experiments, explicitly conditioning the decoding policy on the available sampling budget—and training it across a range of budgets—consistently improves performance, even when evaluated at a fixed budget (Section~\ref{sec:seq-results}). This suggests that exposing the policy to inference-time heterogeneity during training leads to more robust decoding behavior.

Motivated by these gaps, we explore a different axis of inference control:
learning decoding-time policies that adapt sampling strategy based on model
state and available compute budget. Our intuition is informed by recent
observations on so-called “forking tokens” in reasoning
tasks~\cite{wang2025beyond}, which suggest that a small number of high-entropy
decisions can disproportionately influence the outcome of multi-step solutions.
Rather than explicitly branching or performing tree search, we view these
observations as motivation for allowing different parts of a single reasoning
trajectory to exhibit different degrees of stochasticity. In this view,
encouraging exploration at uncertain decision points, while decoding more
deterministically elsewhere, may improve accuracy under fixed compute
constraints. Instead of hand-designing entropy thresholds or heuristics, we seek
to learn such behavior directly from task reward.

To this end, we introduce \emph{Learned Decoding Adapters}, a family of
reinforcement-learning–based policies that modulate decoding during inference
while leaving the underlying language model fixed.  At the sequence level, we
formulate decoding strategy selection as a contextual bandit: the adapter
selects a decoding configuration (e.g.\ greedy, top-$k$, top-$p$, or min-$p$)
for each prompt, conditioned on prompt features and a parallel-sampling budget.
The action space in this setting is constructed via a data-driven greedy
selection procedure over candidate decoding strategies. At the token level, we
treat decoding as a partially observable Markov decision process (POMDP): the
adapter observes internal model representations and the remaining token budget,
and selects a decoding action at each step. While the framework supports
arbitrary decoding configurations, in our experiments we focus on
temperature-based actions for token-level control, which provide a simple and
interpretable axis for dynamically allocating stochasticity within a single
generation trajectory.

We train both adapters using policy-gradient reinforcement learning
(REINFORCE~\cite{williams1992simple}) with verifiable terminal rewards, such as
correctness checks on math and code problems. Experiments on the MATH and
CodeContests benchmarks show that the learned adapters improve the
accuracy-budget tradeoff. The token-level adapter improves Pass@1 accuracy by up
to 10.2\% under fixed token budgets, while the sequence-level adapter
outperforms strong fixed strategies under limited parallel sampling. Ablation
analyses support the complementary contributions of sequence-level and
token-level adaptation.

\paragraph{Contributions.}
We make the following contributions:
\begin{itemize}
    \item We formulate decoding-time inference as a policy learning problem,
    introducing a unified reinforcement learning framework for both prompt-level
    and token-level adaptation under explicit compute budgets.
    \item We propose decoding adapters trained solely with online verifiable
    task rewards—without learned reward models, preference labels, or
    hand-designed decoding heuristics—while keeping the underlying language
    model fixed.
    \item We demonstrate empirical gains on mathematical and coding reasoning
    benchmarks under constrained compute, and analyze how the learned adapters
    allocate stochasticity to improve solution accuracy.
\end{itemize}

\begin{figure}[h] \centering 
\begin{subfigure}[b]{0.49\textwidth} \centering \includegraphics[width=\textwidth]{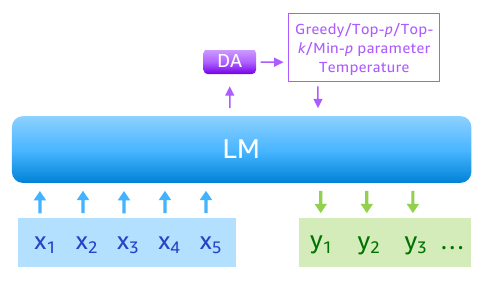} \caption{ \textbf{Sequence-level adapter}: a single DA predicts one decoding configuration that is applied throughout generation.} \label{fig: seq} \end{subfigure} \hfill \begin{subfigure}[b]{0.49\textwidth} \centering \includegraphics[width=\textwidth]{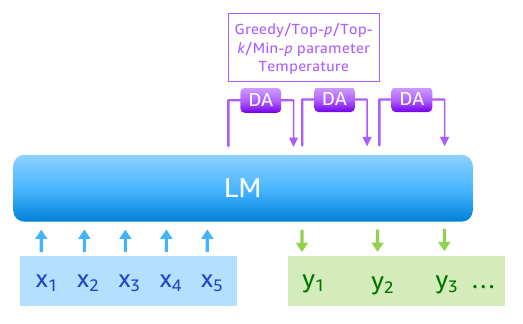} \caption{ \textbf{Token-level adapter}: the DA is invoked at each decoding step to select a (potentially different) decoding configuration per token. } \label{fig: tok} \end{subfigure} \caption{Overview of the proposed decoding adapter (DA) for a frozen language model (LM). Blue blocks denote input tokens \( x_i \); green blocks denote generated tokens \( y_i \). } \label{fig:scheme} \end{figure}

Figure~\ref{fig:scheme} provides an overview of the proposed decoding adapter
framework, illustrating both the sequence-level and token-level policies layered
on top of a frozen language model under explicit compute budgets.

\section{Methods}

\subsection{Preliminaries}

We study inference-time control of decoding for a frozen large language model
(LLM) $f$ under explicit compute budgets. Our goal is to learn a lightweight
\emph{decoding adapter} that modulates how tokens are sampled from $f$ while
leaving all LLM parameters fixed.

\paragraph{Budgets.}
For each problem instance, let $B$ denote a \emph{parallel sampling budget},
quantified as the maximum number of full decoding trajectories that may be
generated, and let $b$ denote a \emph{token budget}, quantified as the maximum
number of decoding steps within a single trajectory. For token-level control, we
also write $b_t = b - t$ for the remaining token budget at decoding step $t$.

\paragraph{Base model distribution and decoding actions.}
Let $q$ denote an input prompt and $y_{<t}$ a partial generation. The frozen LLM
produces next-token logits
$z_t = f(q, y_{<t}) \in \mathbb{R}^{|\mathcal{V}|}$,
inducing a base distribution $p_f(\cdot \mid q, y_{<t}) = \mathrm{Softmax}(z_t)$. Decoding corresponds to sampling from this
distribution, potentially after applying transformations such as temperature
scaling or truncation.

At each decision point (either once per sequence or once per token), the adapter
selects an action from a discrete action space
\[
S = \{ a_1, a_2, \dots, a_m \},
\]
where each action corresponds to a decoding configuration specified by sampling
parameters such as \texttt{temperature}, \texttt{top\_k}, \texttt{top\_p}, and
\texttt{min\_p}. We view each action $a \in S$ as specifying a transformation
$T_a$ applied to the base distribution, yielding an action-conditioned sampling
distribution
\[
p_{f,a}(\cdot \mid q, y_{<t}) \triangleq T_a\!\left(p_f(\cdot \mid q, y_{<t})\right),
\]
from which the next token is sampled. Conceptually, the frozen language model together with a chosen decoding action
induces a stochastic transition kernel over tokens and hidden states. The
decoding adapter operates by selecting among these induced transition dynamics
through its choice of decoding action.

\paragraph{Reward and policy.}
We assume access to a \emph{verifiable terminal reward} function $r$, which
indicates success or failure of the final generated output (e.g.\ correctness on
math or coding problems). The adapter is parameterized by a policy $\pi_\theta(a
\mid x)$, where the input $x$ consists of features derived from the frozen LLM
(e.g.\ prompt embeddings or hidden-state representations derived from the frozen LLM)
. In the
\emph{budget-aware} setting, we additionally include the relevant budget in the policy input:
\[
x = [e; B] \quad \text{or} \quad x_t = [e_t; b_t],
\]
depending on whether decisions are made at the sequence or token level. This
reflects that $B$ governs prompt-level allocation across parallel rollouts,
while $b_t$ governs within-trajectory allocation across decoding steps.






\subsection{Sequence-Level: Contextual Bandits}

In the sequence-level setting, the decoding adapter selects a single decoding
configuration that is applied uniformly within each generation process. Because this decision
is made once per prompt and determines the sampling dynamics for the entire
rollout before a terminal reward is observed, the problem naturally admits a
contextual bandit formulation.

Concretely, each problem instance defines a \emph{context} consisting of a
prompt representation and an inference budget. Given a prompt embedding $e$
produced by the frozen language model and a parallel sampling budget $B$, the
policy
\[
\pi_\theta(a \mid x), \quad x = [e; B],
\]
selects a decoding action $a \in S$. This action specifies a sampling strategy
that is held fixed for the entire generation. The language model then produces a
complete output sequence under the sampling dynamics induced by $a$, after which
a terminal reward $r(x,a)$ is observed.

The inclusion of $B$ in the context reflects that sequence-level decisions
govern how compute is allocated across parallel rollouts. For example, under a
larger parallel sampling budget, more exploratory decoding strategies may be
preferred, whereas smaller budgets favor conservative strategies. In contrast,
sequence-level adapters do not observe token-level progress and therefore
operate at the granularity of whole-sequence allocation.

We train the decoding adapter to maximize expected terminal reward under
the selected decoding policy, while encouraging exploration across
decoding strategies via entropy regularization. Here, the terminal reward $r(x,a)$ corresponds to the evaluation metric induced by
the inference procedure under budget $B$ (e.g., Pass@$k$ or best-of-$B$),
aggregating outcomes across the available parallel samples.
 Formally, the objective
is
\begin{equation}
\mathcal{J}_{\text{seq}}(\theta)
=
\mathbb{E}_{x \sim D,\, a \sim \pi_\theta(\cdot \mid x)}
\bigl[r(x,a)\bigr]
+
\beta \,
\mathbb{E}_{x \sim D}
\bigl[\mathcal{H}(\pi_\theta(\cdot \mid x))\bigr],
\label{eq:seq_objective}
\end{equation}
where $\mathcal{H}$ denotes the entropy of the policy distribution.

In practice, we optimize this objective using a Monte Carlo policy-gradient
estimator with a variance-reducing baseline. We defer implementation
details to the experimental section.







\subsection{Token-Level: Reinforcement Learning} \label{sec:tok_method}

In the token-level setting, the decoding adapter selects actions at each
decoding step, allowing the sampling strategy to vary within a single generation
trajectory. Unlike the sequence-level case, decisions are made repeatedly over
time and influence future states, making this setting naturally modeled as a
partially observable Markov decision process (POMDP). Partial observability
arises because the adapter does not observe the full environment state—only a
compressed representation of the model’s internal activations and the remaining
budget.

Let $b$ denote the maximum token budget for a trajectory, and let
\[
b_t = b - t
\]
be the remaining token budget at decoding step $t$. At each step, the policy observes a compact state representation
\[
x_t = [e_t; b_t],
\]
where $e_t$ is derived from the frozen language model’s hidden state embedding at step $t$. The policy then samples an action
$a_t \sim \pi_\theta(\cdot \mid x_t)$,
which specifies the decoding configuration used to generate the next token.

Including $b_t$ in the policy input reflects that token-level decisions govern how stochasticity is allocated \emph{within} a single trajectory. A natural design intuition is that, when substantial budget remains, exploratory sampling may be beneficial, whereas near the end of a trajectory, more deterministic decoding may help reduce variance and stabilize completion. This contrasts with the sequence-level setting, where the budget governs allocation across parallel rollouts rather than across time.

Given a sequence of actions $\mathbf{a} = (a_1, \dots, a_T)$, the decoding process produces a complete output sequence, after which a terminal reward $r(x, \mathbf{a})$ is observed. The objective is to maximize expected terminal reward:
\begin{equation}
\mathcal{J}_{\text{tok}}(\theta)
=
\frac{1}{N} \sum_{i=1}^N
\mathbb{E}_{\mathbf{a} \sim \pi_\theta}
\bigl[r(x^{(i)}, \mathbf{a})\bigr].
\label{eq:tok_objective}
\end{equation}

We optimize this objective using standard policy-gradient methods with entropy regularization to encourage exploration over decoding actions. As in the sequence-level case, we use a variance-reducing baseline and defer implementation details to the experimental section.

\paragraph{Training stability.}
In practice, naive application of token-level REINFORCE led to high-variance gradients and unstable training.
To mitigate this, we apply two simple but important stabilizations.
First, we filter the training distribution to exclude prompts that produce extremely sparse or noisy reward signals.
Second, we mask tokens whose next-token distribution is already highly concentrated (maximum probability exceeding 0.95),
as these contribute little learning signal while significantly increasing gradient variance.
Without these adjustments, we were unable to obtain stable training for token-level policies.

\subsection{Selection of Action Space}\label{sec:act_select}

While our framework allows for arbitrary decoding actions, in practice we
restrict the action space to a finite set of representative sampling strategies.
This selection yields a compact set of decoding behaviors that are both diverse
and competitive, providing sufficient expressive power to outperform strong
static baselines while keeping the individual decoding actions interpretable.

\paragraph{Sequence-level action selection.}
For the sequence-level adapter, which selects a single decoding strategy per
prompt, we construct the action space using a principled, data-driven selection
procedure inspired by the coverage-based strategy selection approach of
AuPair~\cite{mavalankar2025aupair}. Specifically, we begin with a large candidate pool of decoding configurations
formed by combinations of temperature, top-$k$, top-$p$, and min-$p$ values (see
\Cref{appendix:action_selection}). The candidate pool is designed to span a range
of commonly used decoding behaviors, from near-greedy to highly stochastic
sampling. While these parameters can have overlapping effects, jointly varying
them enables the selection procedure to identify qualitatively distinct
strategies that perform well on different subsets of inputs. We then evaluate these configurations on held-out validation data and select a
small subset of strategies that together provide strong coverage of
high-performing behaviors.

Concretely, the goal is to select a set of actions $S \subseteq \mathcal{C}$ such
that, across inputs, the best-performing strategy within $S$ achieves high
reward. Intuitively, this encourages diversity among the selected actions while
avoiding redundant configurations. This selection can be viewed as a greedy
approximation to maximizing the expected performance of the induced
``best-of-$S$'' decoder, analogous to submodular maximization objectives used in
prior work~\cite{mavalankar2025aupair}.

Formally, let $\mathcal{C}$ denote the candidate pool and define the set function
\[
F(S) \triangleq \sum_{x \in \mathcal{D}_{\mathrm{val}}} \max_{s \in S} R(x,s),
\]
where $R(x,s)$ is the reward obtained on instance $x$ under strategy $s$.
We choose $S \subseteq \mathcal{C}$ with $|S|=k$ by approximately solving
\begin{equation}
\max_{S \subseteq \mathcal{C}:\ |S|=k} F(S).
\label{eq:action_selection_objective}
\end{equation}
This objective is a standard instance of monotone submodular maximization, for
which greedy selection provides a constant-factor
approximation~\cite{nemhauser1978analysis}. For completeness, we provide a full
description of this procedure in Appendix~\ref{appendix:action_selection}.

\paragraph{Token-level action space.}
For the token-level adapter, we adopt a more restricted action space and focus
on temperature-based decoding actions, while holding other sampling parameters
fixed. Although the framework supports arbitrary decoding configurations at each
step, we found temperature to be a particularly effective and interpretable axis
for token-level control. In contrast, dynamically varying truncation-based
parameters such as top-$k$ or top-$p$ at the token level introduces additional
complexity in interpretability, without yielding qualitatively different
behavior in our setting.

Empirically, this restriction did not limit performance and enabled clearer
insight into how the learned policy allocates stochasticity over time within a
single trajectory. Importantly, this choice reflects an experimental design
decision rather than a limitation of the framework, which remains agnostic to
the specific parameterization of decoding actions.
\section{Sequence-Level Experiments}
\label{sec:seq}

In this section, we evaluate the \emph{sequence-level} decoding adapter, which selects a single decoding strategy per prompt under an explicit parallel sampling budget. We focus on two representative reasoning domains: mathematical problem solving and competitive programming. We set the max output length to 8092 for all sequence- and token-level experiments, except for AIME'25, where we extend this length to 38,912 as in \cite{qwen3technicalreport}.

\subsection{Experimental Setup}
\label{sec:seq-setup}

\paragraph{Datasets.}
We conduct experiments on MATH~\citep{hendrycksmath2021} and CodeContests~\citep{li2022alphacode}. MATH consists of competition-style math problems requiring multi-step reasoning, while CodeContests contains algorithmic programming problems. Both benchmarks exhibit substantial heterogeneity in problem structure and uncertainty, making them well-suited for evaluating adaptive decoding.

\paragraph{Models.}
We primarily report results using Qwen3-4B~\citep{qwen3technicalreport}, which offers a favorable trade-off between model size and reasoning performance and supports both ``thinking'' and ``non-thinking'' generation modes. Additional results on Qwen2.5-Math-1.5B~\citep{yang2024qwen25mathtechnicalreportmathematical} and Qwen3-8B are provided in the appendix.

\paragraph{Decoding strategies and baselines.}
We first construct a candidate pool of decoding configurations and greedily select a small action set separately for MATH, CodeContests, and a mixed math+code validation set. The selection procedure is described in \Cref{appendix:action_selection}. We compare against two strong static baselines: (\textsc{Best}), the single fixed strategy that achieves the highest validation performance on the evaluation dataset, and (\textsc{Mixed}), a uniform mixture over the selected action set. The latter uses the same action space as the adapter but without learning.

\paragraph{Prompt distributions and inference budgets.}
We evaluate under both standard prompting and mixed Chain-of-Thought (CoT) prompting. We evaluate under both standard prompting and mixed Chain-of-Thought (CoT) prompting.
Mixed CoT prompting is used to deliberately induce heterogeneity in reasoning styles during training, ensuring that the adapter does not overfit to a single prompt format and remains robust to inference-time variation.
For CoT-mixed training, a fixed fraction of prompts is randomly augmented with a CoT template, while the remainder use direct-answer prompting. We also evaluate multiple inference strategies, including Pass@$k$ with $k \in \{1,2,4,8\}$, which capture different tradeoffs between compute and reliability. The parallel sampling budget is provided as part of the policy input when training budget-aware variants. We encode the budget value using a two-layer MLP, then concatenate it to the embedding of the context.

\subsection{Main Results}
\label{sec:seq-results}

We report mean accuracy with 95\% confidence intervals, aggregated over $k{=}3$ independent runs.

\paragraph{Single-domain results.}
Tables~\ref{table:seq-level-results} (MATH) and ~\ref{table:seq-level-coding-results} (CodeContests) summarize performance of the sequence-level adapter versus static baselines. Across all settings, the learned adapter outperforms both \textsc{Best} and \textsc{Mixed}, with gains increasing under larger parallel sampling budgets, indicating improved use of additional rollouts. From \Cref{fig:length-distribution}, we see improved performance is \textbf{not} due to extended length.

\paragraph{Effect of budget conditioning and mixed-strategy training.}
Conditioning the policy on the parallel sampling budget consistently improves performance relative to budget-agnostic training. Training on a mixture of inference strategies further improves robustness to evaluation-time heterogeneity, with the largest gains observed under mixed CoT prompting. Together, these results indicate that explicitly modeling inference-time constraints is a key driver of the improvements.

\begin{table*}[ht]
\centering
\caption{\textbf{MATH.} Comparison of static sampling baselines and the sequence-level adapter under different settings. For \emph{CoT mix}, we apply CoT prompting to 30\% of prompts (ratio $=0.3$). Static sampling reports the best single strategy (\textsc{Best}) and a fixed mixture (\textsc{Mixed}). The adapter is evaluated with a budget provided in feature set and without. Values are percentages (mean $\pm$ confidence interval).}
\label{table:seq-level-results}
\small

\setlength{\tabcolsep}{7pt}
\renewcommand{\arraystretch}{1.1}
\begin{tabular}{ll *{4}{c}}
\toprule
 & & \multicolumn{2}{c}{\bf Static sampling} & \multicolumn{2}{c}{\bf Sequence-level adapter} \\
\cmidrule(lr){3-4}\cmidrule(lr){5-6}
\multicolumn{1}{c}{\bf Metric} & \multicolumn{1}{c}{\bf Setting}
& \multicolumn{1}{c}{ Best }
& \multicolumn{1}{c}{\makecell{Mixed}}
& \multicolumn{1}{c}{\makecell{w/o budget}}
& \multicolumn{1}{c}{\makecell{w/ budget}} \\
\midrule
\multirow{5}{*}{Pass@1}
 & TURN~\cite{du2025optimizing} & \multicolumn{4}{c}{ 72.72 } \\
 \cmidrule(lr){2-6}
 & w/o CoT & 71.70 \(\pm\) 1.25 & 71.20 \(\pm\) 1.26 & 72.60 \(\pm\) 1.24  & \bestcell{ 72.90 \(\pm\) 1.23 } \\
 
 & \(\Delta\) Abs. (Rel.) &  &  & \textcolor{darkgreen}{$\uparrow$ 0.90 (+1.26\%)} & \textcolor{darkgreen}{$\uparrow$1.20 (+1.67\%)} \\
 
 & mix CoT & 72.10 \(\pm\) 1.24 & 71.93 \(\pm\) 1.25 &  73.60 \(\pm\) 1.22 & \bestcell{ \textbf{74.20} \(\pm\) 1.21 } \\
 
 & \(\Delta\) Abs. (Rel.) &  & & \textcolor{darkgreen}{$\uparrow$1.50 (+2.08\%)} & \textcolor{darkgreen}{$\uparrow$2.10 (+2.91\%)} \\
\addlinespace
\cmidrule(lr){1-6}
\multirow{5}{*}{\makecell{Pass@8}}
& TURN~\cite{du2025optimizing} & \multicolumn{4}{c}{ 76.96 } \\
\cmidrule(lr){2-6}
 & w/o CoT &  76.70 \(\pm\) 1.17 & 76.23 \(\pm\) 1.18 &  78.30 \(\pm\) 1.14  &  \bestcell{ 78.46 \(\pm\) 1.14 } \\
 & \(\Delta\) Abs. (Rel.) &  &  & \textcolor{darkgreen}{$\uparrow$ 1.60 (+2.09\%)} & \textcolor{darkgreen}{$\uparrow$ 1.76 (+2.29\%)} \\
 & mix CoT &  77.10 \(\pm\) 1.16 & 76.57 \(\pm\) 1.17  &  78.80 \(\pm\) 1.13  &  \bestcell{ \textbf{79.80} \(\pm\) 1.11 } \\
 & \(\Delta\) Abs. (Rel.) &  &  & \textcolor{darkgreen}{$\uparrow$1.70 (+2.20\%)} & \textcolor{darkgreen}{$\uparrow$ 2.70 (+3.50\%)} \\
 \addlinespace
\bottomrule
\end{tabular}
\end{table*}

\begin{table*}[ht]
\centering
\caption{\textbf{CodeContests}: comparison of static sampling strategies vs. the \textbf{sequence}-level sampling adapter under different settings.}
\label{table:seq-level-coding-results}
\small
\setlength{\tabcolsep}{7pt}
\renewcommand{\arraystretch}{1.1}
\begin{tabular}{ll *{4}{c}}
\toprule
 & & \multicolumn{2}{c}{\bf Static sampling} & \multicolumn{2}{c}{\bf Sequence-level adapter} \\
\cmidrule(lr){3-4}\cmidrule(lr){5-6}
\multicolumn{1}{c}{\bf Metric} & \multicolumn{1}{c}{\bf Setting}
& \multicolumn{1}{c}{Best}
& \multicolumn{1}{c}{\makecell{Mixed}}
& \multicolumn{1}{c}{\makecell{ w/o budget }}
& \multicolumn{1}{c}{\makecell{w/ budget}}
\\
\midrule
\multirow{5}{*}{Pass@1}
 & w/o CoT  & 11.43 \(\pm\) 2.28 & 10.53 \(\pm\) 2.19 & \bestcell{ 14.53 \(\pm\) 2.52 } &  14.50 \(\pm\) 2.52  \\
 
 & \(\Delta\) Abs. (Rel.) &  &  & \textcolor{darkgreen}{$\uparrow$ 3.10
 (+27.12\%)} & \textcolor{darkgreen}{$\uparrow$ 3.07 (+26.85\%)} \\
 
 & mix CoT & 13.97 \(\pm\) 2.48 & 14.80 \(\pm\) 2.54 & 17.06 \(\pm\) 2.69  & \bestcell{ \textbf{19.70} \(\pm\) 2.85 } \\
 
 & \(\Delta\) Abs. (Rel.) &  &  & \textcolor{darkgreen}{$\uparrow$ 2.26
 (+15.27\%)} & \textcolor{darkgreen}{$\uparrow$ 4.90 (+33.11\%)} \\
 \addlinespace
 \cmidrule(lr){1-6}
\multirow{5}{*}{Pass@8}
 & w/o CoT & 22.80 \(\pm\) 3.00 & 22.23 \(\pm\) 2.97 & \bestcell{ 26.08 \(\pm\) 3.14 } &  23.10 \(\pm\) 3.02 \\
 & \(\Delta\) Abs. (Rel.) &  &  & \textcolor{darkgreen}{$\uparrow$ 3.28 (+14.39\%)} & \textcolor{darkgreen}{$\uparrow$ 0.30 (+1.32\%)} \\
 & mix CoT  & 29.10 \(\pm\) 3.25 & 25.63 \(\pm\) 3.12 & 29.90  \(\pm\) 3.28 & \bestcell{ \textbf{32.50} \(\pm\) 3.35   }   \\
 & \(\Delta\) Abs. (Rel.) &  &  & \textcolor{darkgreen}{$\uparrow$ 0.80 (+2.75\%)} & \textcolor{darkgreen}{$\uparrow$ 3.40 (+11.68\%)} \\
\addlinespace
\bottomrule
\end{tabular}
\end{table*}

\subsection{Mixed-Domain Training}
\label{sec:seq-mixed}

Table~\ref{table:seq-mixed} reports results when training a single sequence-level adapter jointly on math and coding data (with balanced sampling during training) and evaluating on each domain’s original validation distribution. The adapter continues to improve over static baselines on both domains, though gains are smaller than in single-domain training. This reduction is consistent with the increased diversity of the joint task distribution, and highlights that the adapter can learn compromise decoding policies under heterogeneous workloads.

\subsection{Policy Behavior and Qualitative Analysis}
\label{sec:seq-policy}

\begin{figure}[h]
\centering
    \centering
    \includegraphics[width=0.5\textwidth]{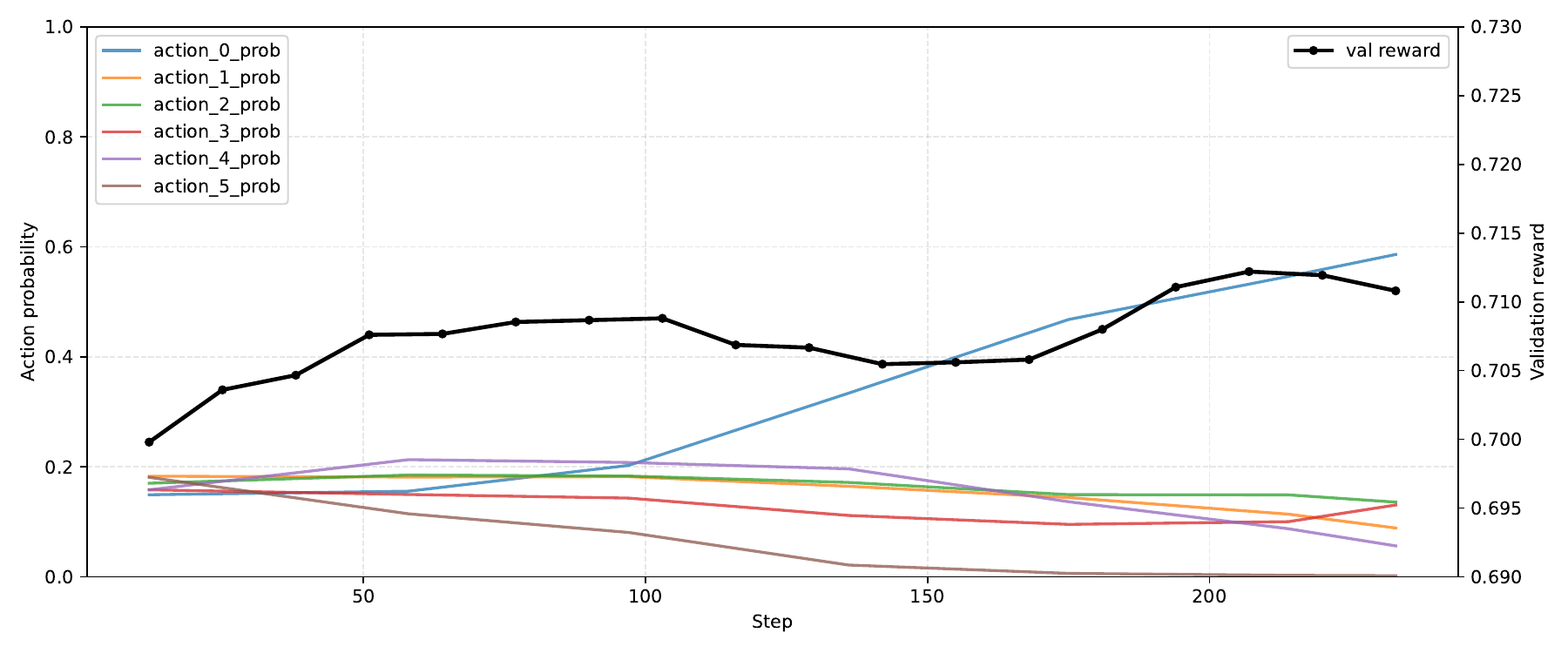}
%
\caption{\textbf{Action distributions and validation reward on MATH} for sequence-level adapter strategies. Actions are defined in \cref{tab:math}. The x-axis denotes training progress; the left y-axis reports action probability; the right y-axis reports validation reward.}
\label{fig:math-action-distribution}
\end{figure}

\paragraph{Policy behavior.}
Figures~\ref{fig:math-action-distribution} and~\ref{fig:coding-action-distribution} illustrate how the sequence-level adapter allocates probability mass over decoding strategies during training. Rather than collapsing to a single strategy, the learned policy concentrates mass on a small subset of high-performing actions while maintaining non-zero probability on alternatives. This behavior is consistent with a learned trade-off between exploitation and robustness: the adapter favors reliable strategies under tight budgets, while preserving optionality when additional parallel samples are available.

\paragraph{Prompt-type and metric stability.}
At the sequence level, action distributions are broadly consistent between CoT and non-CoT prompts, suggesting that optimal strategy selection is driven primarily by context and budget considerations rather than by the presence of CoT prompts. We also observe only minor differences in preferred strategies across inference metrics (e.g., Pass@1 vs.\ Pass@8), indicating that the learned policy generalizes across evaluation criteria rather than overfitting to a single inference objective. In contrast, prompt-dependent differences are more pronounced in token-level adaptation, where fine-grained uncertainty signals become salient.

\subsection{Generalization}
\label{sec:seq-generalization}

To assess out-of-domain generalization, we evaluate the adapter trained on MATH on (i) the CodeContests evaluation set and (ii) the harder mathematical reasoning dataset AIME 2025~\citep{AIME}. No additional tuning is performed on AIME.
Since AIME'25 is small, we run 30 random seeds. Results in \Cref{table:aime} and \Cref{table:seq-gen-to-coding} show that the MATH-trained sequence-level adapter remains competitive and can improve over reported baselines, supporting the view that the learned strategy-selection policy captures transferable signals beyond the training distribution.

\begin{table}[t]
\centering
\caption{AIME'25 performance comparison of a \textbf{sequence}-level adapter trained on MATH-train vs.\ numbers reported in the Qwen3-4B technical report~\citep{qwen3technicalreport}.}
\label{table:aime}
\small
\setlength{\tabcolsep}{7pt}
\begin{tabular}{ll *{4}{c}}
\toprule
\textbf{Metric} & \textbf{Setting} & \multicolumn{1}{c}{ \textbf{Reported}} & \multicolumn{1}{c}{ \textbf{Adapter}  (w/ budget)} \\
\midrule
\multirow{2}{*}{Pass@1} & non-thinking & 19.1 & 20.1 \(\pm\) 2.62 \\
                        & thinking     & 65.6 & 71.1 \(\pm\) 2.96 \\
\bottomrule
\end{tabular}
\end{table}

\begin{table*}[ht]
\centering
\caption{Mixed training of math and coding: comparison of static sampling strategies vs. the \textbf{sequence}-level sampling adapter. We use \textbf{Code} to refer to CodeContests in this table.}
\label{table:seq-mixed}
\small
\setlength{\tabcolsep}{1.5pt}
\begin{tabular}{ll *{8}{c}}
\toprule
 & & \multicolumn{4}{c}{\bf Static sampling} & \multicolumn{4}{c}{\bf Sequence-level adapter} \\
\cmidrule(lr){3-6}\cmidrule(lr){7-10}
\multicolumn{1}{c}{\bf Metric} & \multicolumn{1}{c}{\bf Setting}
& \multicolumn{2}{c}{Best}
& \multicolumn{2}{c}{Mixed}
& \multicolumn{2}{c}{w/o  budget}
& \multicolumn{2}{c}{\makecell{w/ budget}} \\
\cmidrule(lr){3-4}\cmidrule(lr){5-6}\cmidrule(lr){7-8}\cmidrule(lr){9-10}
 &  & \textbf{MATH} & \textbf{Code} & \textbf{MATH} & \textbf{Code} & \textbf{MATH} & \textbf{Code} & \textbf{MATH} & \textbf{Code} \\
\midrule
\multirow{2}{*}{Pass@1}
 & w/o CoT 
   & 72.00 \(\pm\) 3.18 & 11.80 \(\pm\) 2.39 & 71.10 \(\pm\) 3.17 & 10.30 \(\pm\) 2.18 & 73.55 \(\pm\) 3.12 & 13.70 \(\pm\) 2.46 & \bestcell{ \textbf{74.20} \(\pm\) 3.13 } & \bestcell{ 13.70 \(\pm\) 2.46 } \\
   
   & \(\Delta\) Abs. (Rel.) & & & & & \textcolor{darkgreen}{$\uparrow$ 1.55
 (2.15\%)} & \textcolor{darkgreen}{$\uparrow$ 1.90
 (16.10\%)} & \textcolor{darkgreen}{$\uparrow$ 2.20 (3.06\%)} & \textcolor{darkgreen}{$\uparrow$ 1.90 (16.10\%)} \\
 & mix CoT 
   & 73.80 \(\pm\) 3.15 & 17.10 \(\pm\) 2.69 & 73.30 \(\pm\) 3.17  & 14.50 \(\pm\) 2.52 & \bestcell{ 74.80 \(\pm\) 3.11 } & \bestcell{ \textbf{18.50}\(\pm\) 2.89 } &  74.70 \(\pm\) 3.11 &  18.20 \(\pm\) 2.76\\
   
   & \(\Delta\) Abs. (Rel.) & & & & & \textcolor{darkgreen}{$\uparrow$1.00 (1.36\%)} & \textcolor{darkgreen}{$\uparrow$ 1.40 (8.19\%)} & \textcolor{darkgreen}{$\uparrow$ 0.90 (1.22\%)} & \textcolor{darkgreen}{$\uparrow$ 1.10 (6.43\%)} \\
\addlinespace
\cmidrule(lr){1-10}
\multirow{4}{*}{Pass@8}
 & w/o CoT 
   & 76.30 \(\pm\) 3.04 & 14.50 \(\pm\) 2.52 & 75.70 \(\pm\) 3.07 & 14.50 \(\pm\) 2.52
   & 77.70 \(\pm\) 2.97 &   17.90 \(\pm\) 2.74 
   & \bestcell{ 78.00 \(\pm\) 2.96 } & \bestcell{ 19.70 \(\pm\) 2.85 } \\
 & \(\Delta\) Abs. (Rel.) & & & & & \textcolor{darkgreen}{$\uparrow$ 0.70 (0.92\%)} & \textcolor{darkgreen}{$\uparrow$ 4.50 (31.03\%)} & \textcolor{darkgreen}{$\uparrow$ 1.70 (2.23\%)} & \textcolor{darkgreen}{$\uparrow$ 5.20 (35.86\%)} \\
 & mix CoT 
   & 76.70 \(\pm\) 2.98 & 22.20 \(\pm\) 2.97 & 77.10 \(\pm\) 2.98 &  18.80 \(\pm\) 2.80  & 77.70 \(\pm\) 2.93 & \bestcell{ \textbf{24.80} \(\pm\) 3.09 }
   & \bestcell{ \textbf{78.10} \(\pm\) 2.96 } &  23.10 \(\pm\) 3.02 \\
   & \(\Delta\) Abs. (Rel.) &  &  &  &  & \textcolor{darkgreen}{$\uparrow$ 0.60 (0.78\%)} & \textcolor{darkgreen}{$\uparrow$ 2.60 (11.71\%)} & \textcolor{darkgreen}{$\uparrow$ 1.00 (1.30\%)} & \textcolor{darkgreen}{$\uparrow$ 0.90 (4.05\%)} \\
\bottomrule
\end{tabular}
\end{table*}

\section{Token-Level Experiments}

We evaluate the token-level decoding adapter, which selects a decoding action at
each generation step, allowing stochasticity to vary within a single trajectory.
This enables finer-grained inference-time control than the sequence-level adapter,
which commits to a single decoding strategy per prompt. As in the sequence-level
setting, we compare against strong static sampling baselines. To isolate the effect
of token-level adaptation, we restrict the action space to four temperature-based
decoding strategies (Table~\ref{tab:math-token}).

\paragraph{Setup.}
We train token-level adapters under a fixed per-sequence token budget and consider
two variants: one that conditions only on the usual model context, and one that
additionally conditions on the remaining token budget. Both variants are trained
using the same reinforcement-learning objective described in
Section~\ref{sec:tok_method}.

For static baselines, we evaluate greedy decoding and a fixed mixture over the same
four temperature actions used by the adapter. The latter uses an identical action
space but does not condition on context or budget. We note that the action space
used here differs from the richer set of decoding strategies considered in the
sequence-level experiments; within this restricted temperature-only space, greedy
decoding achieves the strongest static performance on validation without chain of though and therefore serves as the primary static baseline.

\paragraph{Main results.}
Results on MATH are summarized in Table~\ref{table:tok-result}. The token-level
adapter substantially outperforms all static decoding strategies. Conditioning on
the remaining token budget yields the strongest performance, improving Pass@1 by
approximately 9--10\% over the best static baseline. Even without explicit budget
conditioning, the token-level adapter achieves consistent gains over static
decoding, indicating that per-token adaptation alone provides meaningful benefits.
\begin{table*}[ht]
\centering
\caption{\textbf{Comparison of sampling strategies on Math}: performance comparison of \textbf{token}-level sampling adapter versus static sampling strategies. Here \texttt{budget} is the remaining token budget within a sequence. Values are percentages.}
\label{table:tok-result}
\small
\begin{tabular}{ll *{6}{c}}
\toprule
 & & \multicolumn{2}{c}{\bf Static sampling} & \multicolumn{4}{c}{\bf Adapter} \\
\cmidrule(lr){3-4}\cmidrule(lr){5-8}
\multicolumn{1}{c}{\bf Metric} & \multicolumn{1}{c}{\bf Setting}
& \multicolumn{1}{c}{Greedy}
& \multicolumn{1}{c}{\makecell{ Mixed}}
& \multicolumn{1}{c}{w/o budget}
& \multicolumn{1}{c}{\(\Delta\)}
& \multicolumn{1}{c}{budget} 
& \multicolumn{1}{c}{\(\Delta\)} 
\\
\midrule
\multirow{2}{*}{Pass@1} & w/o CoT & 71.33 \(\pm\) 1.25 & 71.60 \(\pm\) 1.25 & 78.28 \(\pm\) 1.14 & +6.68 & \bestcell{ 80.82 \(\pm\) 1.07 } & +9.49 \\
 & mix CoT & 72.10 \(\pm\) 1.24 & 72.67 \(\pm\) 1.24 & 78.52 \(\pm\) 1.14 & +5.85 & \bestcell{ \textbf{82.33} \(\pm\) 1.03 } & +10.23 \\
\bottomrule
\end{tabular}
\end{table*}

Notably, the magnitude of these improvements exceeds those observed for
sequence-level adaptation under comparable evaluation settings, highlighting the
value of fine-grained, within-trajectory control.



\paragraph{Is entropy alone sufficient?}
To test whether token-level adaptation can be reduced to entropy-based heuristics,
we train an ablation in which the adapter observes only token entropy rather than
the full contextual representation. This entropy-only policy reaches a validation
reward of 72.03 (averaged over 3 seeds), comparable to static decoding
baselines but far below the full token-level adapter, which exceeds 80\% Pass@1.
This indicates that while entropy correlates with uncertainty, entropy alone is
insufficient to recover the gains achieved by learned token-level adaptation.

\paragraph{Qualitative analysis.}
To better understand the behavior of the learned token-level policy, we conducted
additional qualitative analyses examining token entropy statistics and token
bucketing by entropy, position, and related summary measures
(Appendix~\ref{app:qual}). We observe that lower-entropy tokens are more frequently
collapsed to near-deterministic behavior under the learned policy, while
higher-entropy tokens tend to retain greater stochasticity. However, despite this
trend, we did not identify simple qualitative rules or low-dimensional heuristics
that fully explain the policy’s behavior. This suggests that the gains from
token-level adaptation are not attributable to a small number of easily
interpretable patterns.

\paragraph{Policy behavior.}
Figure~\ref{fig:token-level-action-distribution} shows the evolution of action
selection during training. Across settings, the learned policy concentrates most
probability mass on the strongest static strategy while maintaining non-zero
probability on alternatives. Validation reward increases as the policy converges
toward this distribution. Because a fully collapsed policy would reduce to static
decoding, these results indicate that improvements arise from selective,
context-dependent deviations rather than wholesale changes in decoding behavior.


\section{Related Work}

\paragraph{Adaptive Decoding Strategies.}
Sampling hyperparameters such as temperature, top-$p$, and top-$k$ are typically fixed across inputs and decoding steps, despite substantial variation in prompt difficulty and uncertainty. Several recent methods seek to adapt these parameters dynamically. AutoDeco \cite{wang2025end} trains lightweight controller heads to select decoding parameters per token. Similarly, \citet{dhuliawala2024adaptive} propose a latent preference optimization framework to adjust generation behavior on the fly. These methods typically rely on learned preferences or reward models. In contrast, our decoding adapters are trained solely with task-level correctness signals and condition explicitly on compute budgets.

\paragraph{Inference-Time Decision Making.}
Other work frames generation as a sequential decision process. Meta-RL approaches \cite{qu2025optimizing} optimize token-level compute use over multi-step reasoning. Entropy-aware objectives \cite{cui2025entropy} preserve exploration during fine-tuning. Unlike these approaches, our decoding policies are trained on frozen models and operate only at inference time. Closely related is \cite{yan2025mur}, which adapts reasoning depth based on uncertainty dynamics, and D-LLM \cite{jiang2024d}, which adjusts computation per token. Our method differs by learning \emph{what decoding strategy to use}, rather than \emph{how much compute to spend}.

\paragraph{Critical Tokens and Adaptive Sampling.}
Recent work identifies “forking” tokens—localized points of uncertainty with high downstream impact \cite{wang2025beyond}—and proposes to adapt generation around them. Our token-level adapter makes such adaptation learnable via reinforcement learning. While some methods (e.g., \cite{zhao2025majority, fu2025deep}) post-process multiple samples to aggregate solutions or early-stop, we focus on decoding-time control without auxiliary reranking.

For additional background on sampling techniques and compute-aware inference, see Appendix~\ref{sec:expanded-related}.

\section{Future Directions}

Our work suggests several directions for extending adaptive decoding beyond the setting studied here.

\paragraph{Richer decoding policies.}
We restrict actions to a small discrete set of decoding operators, but decoding can more generally be viewed as a learnable mapping from the base model distribution to an induced sampling distribution.
Moving to more expressive, continuously parameterized decoding transformations would fundamentally change the learning problem, introducing new challenges in optimization and sample efficiency.

\paragraph{Expanded action spaces.}
The framework naturally supports decoding controls beyond sampling hyperparameters.
For example, incorporating early stopping or adaptive termination would allow joint control over stochasticity and generation length under fixed compute budgets.
More broadly, decoding adapters could allocate sampling effort within structured inference procedures, such as tree search or other multi-trajectory methods, where actions govern both branching and budget allocation.

\paragraph{Joint training with the base model.}
We treat the language model as a frozen environment and learn only a lightweight decoding policy.
An important direction is to study joint or alternating training of the adapter and the underlying model, which may reduce the burden on the decoding policy by shaping representations to better support adaptive inference.

\section{Conclusion}

We framed decoding as a control problem and showed that lightweight policies, trained from scratch with sparse verifiable rewards, can adapt inference behavior under explicit compute budgets.
By learning to select among decoding operators at the sequence and token levels, without modifying the underlying language model,our adapters consistently improve reasoning performance on mathematical and coding tasks.
These results suggest that inference-time control is a meaningful and underexplored axis for improving LLM reasoning, complementary to model scaling and fine-tuning.


\section*{Impact Statement}
This paper presents work whose goal is to advance the field of Machine Learning. There are many potential societal consequences of our work, none which we feel must be specifically highlighted here.


\bibliography{ref}
\bibliographystyle{icml2026}

\newpage
\appendix
\onecolumn
\section{Expanded Related Work}\label{sec:expanded-related}

Large language models (LLMs) have made significant progress in reasoning, code generation, and open-ended tasks, but their decoding behavior at inference time remains a blunt instrument: hyperparameters like temperature or top-$p$ are typically fixed per model or dataset. Recent work has begun to explore decoding-time adaptation, dynamic compute allocation, and reward-supervised control. We review these directions below, emphasizing where our approach of learning decoding strategies under budget constraints via verifiable rewards differs.

\subsection{Sampling Strategies and Their Limitations}
Sampling from the softmax distribution is commonly modified to balance diversity and precision. Top-$k$, top-$p$ (nucleus), temperature scaling, and min-$p$ sampling are widely used, but typically tuned offline and fixed across prompts. Nucleus sampling \citet{holtzman2019curious} avoids low-probability tails, but subsequent work shows its effect on memorization is limited. \citet{borec2024unreasonable} find that increasing the nucleus reduces memorization modestly but does not eliminate it. \citet{schaeffer2025turning} critique the empirical basis for min-$p$ sampling, arguing that evaluation artifacts obscure its true behavior. \citet{grubisic2024priority} propose Priority Sampling, a deterministic variant that avoids duplicate or incoherent outputs caused by stochastic sampling at high temperatures.

These methods reflect the need to modulate decoding behavior, but treat decoding hyperparameters as static knobs. They provide no mechanism for input- or token-specific control, and do not incorporate performance feedback.

\subsection{Adaptive Decoding Architectures}
Several recent works propose models that adapt decoding behavior based on context. AutoDeco \citet{wang2025end} adds lightweight controller heads to predict temperature and top-$p$ at each generation step, trained end-to-end with preference modeling. \citet{dhuliawala2024adaptive} introduce a latent preference optimization (LPO) framework where an LLM selects a discrete preference vector to modulate generation style based on prompt features or learned guidance. \citet{du2025optimizing} improve decoding quality by sampling multiple outputs at different temperatures and selecting the best via reranking, optimizing temperature allocation at the example level.

Learned inference-time controllers have also emerged as a flexible way to steer frozen LLMs without fine-tuning. \citet{dhuliawala2024adaptive}'s AdaptiveDecoder head selects sampling temperature token-by-token using preference-trained RL, outperforming fixed-temperature decoding across tasks. Our approach is similar in spirit, but differs by conditioning on compute budgets and using task-verifiable correctness as reward. Collab \cite{chakraborty2025collab} takes this further by learning a token-level Q-learning policy that selects among expert LLMs mid-generation, effectively creating a mixture-of-agents decoding strategy with substantial quality gains. Cascade Reward Sampling (CARDS) \cite{li2024cascade} guides a frozen model using a learned reward model that scores and filters sampled segments, avoiding low-reward continuations. Controlled decoding methods further bias token choice using learned value functions: \citet{mudgal2023controlled} optimize a prefix-level scorer to guide greedy decoding toward high-reward outcomes, while Integrated Value Guidance \cite{liu2024inference} blends implicit and explicit value heads to align frozen models at inference time.

While these controllers improve alignment or output quality, they typically rely on trained reward models, preference supervision, or post-hoc reranking. In contrast, we train decoding policies tabula rasa from correctness-based rewards, using no learned value models and keeping the LLM frozen. Our approach highlights decoding-time stochasticity as a learnable control surface, and shows that generalization to unseen compute budgets is possible with minimal assumptions.

Moreover, we introduce budget-aware decoding: our policies explicitly condition on token or sampling budget and are trained across a range of compute regimes. This enables generalization to unseen inference-time constraints, a dimension not addressed by prior adaptive decoders.

\subsection{Inference-Time Decision Making and Compute Allocation}
Beyond decoding parameters, recent work explores when and where to allocate compute during generation. \citet{qu2025optimizing} propose meta-RL to optimize token-level compute under regret-based objectives. \citet{cui2025entropy} design entropy-aware policy updates that avoid overconfident collapse in RL fine-tuning.

Several systems use uncertainty signals to trigger additional reasoning. MUR (\citet{yan2025mur}) monitors the momentum of token entropy to decide when to continue or halt reasoning. \citet{jiang2024d} learn to skip transformer layers per token, reducing inference cost without quality degradation.

These methods focus on \emph{when} and \emph{how much} to compute, but not \emph{how to sample}. Our work is orthogonal and complementary: given a token or prompt, we learn how to sample the next generation step, trading off exploration and determinism under budget.

\subsection{Critical Tokens and Token-Level Adaptation}
Multiple studies suggest that reasoning failures often hinge on a small set of high-impact tokens. \citet{wang2025beyond} identify "forking" tokens with high entropy and high downstream influence. \citet{abdin2024phi} define "pivotal tokens" as those with strong local gradients with respect to outcome success.

These findings motivate adaptive sampling that varies across a sequence. AdaDec \cite{he2025adadec} uses uncertainty spikes to trigger beam-style reranking at critical steps. Unlike such heuristics, our token-level adapter is trained via reinforcement learning to make sampling decisions per step. By observing internal model states and remaining token budget, the adapter learns to allocate stochasticity where it matters.

\subsection{Aggregation and Speculative Decoding}
Several methods post-process multiple samples to select better outputs. AggLM \citet{zhao2025majority} learns a solution aggregator via RL, outperforming simple voting on complex reasoning. \citet{fu2025deep} use token probabilities to filter low-quality reasoning paths mid-generation. Speculative decoding \cite{huang2024context, liu2025adaptive} accelerates generation via draft-verification, and some variants use contextual bandits to choose draft strategies.

These methods improve sample efficiency or response quality, but require generating multiple candidates and selecting among them. Our approach is distinct: we aim to improve \emph{per-sample} quality by learning how to generate, not just how to rerank. Our adapters operate during generation, not after it.

\subsection{Fine tuning LLMs with Reinforcement Learning}
Reinforcement learning has emerged as a powerful tool for aligning language models with task objectives or human preferences. In reinforcement learning from human feedback (RLHF), language models are fine-tuned to maximize a learned reward model derived from human preference comparisons, as in InstructGPT \cite{ouyang2022training} and Anthropic's helpful-harmless framework \cite{bai2022training}. Reinforcement learning from verifiable rewards (RLVR) replaces noisy preference models with automated correctness checks, enabling reward supervision from ground-truth signals in domains such as math and code. Group-Relative Policy Optimization (GRPO) \cite{shao2024deepseekmath,guo2025deepseek} further improves efficiency by using group-normalized Monte Carlo advantages, and has shown strong performance on reasoning tasks. These approaches treat the language model itself as the policy, optimized via PPO-style methods. In contrast, our approach treats the language model as the \emph{environment} and learns a lightweight decoding-time policy that interacts with its sampling behavior. This tabula rasa RL framing enables adaptive inference under compute budgets, without modifying the base model.

\subsection{Positioning}
Across these lines of work, few methods explicitly frame decoding as a policy learning problem, or incorporate budget constraints as part of the learning signal. Our formulation unifies these axes: we learn decoding-time policies at both the sequence and token level, trained under verifiable rewards and explicit compute budgets, without modifying the base model.

We view this as a complementary direction to prior work: rather than designing or searching over decoding strategies, we make decoding strategy \emph{learnable}, enabling flexible and efficient adaptation across tasks, budgets, and token-level uncertainties.

\section{Additional experiments}
\label{sec:app}

\subsection{Additional Models}

In the following, we provide additional experiment with two other models: Qwen3-8B and Qwen2.5-Math-1.5B. 

\begin{table*}[ht]
\centering
\caption{\textbf{Comparison of sampling strategies on MATH.}  against the proposed \textbf{sequence}-level sampling adapter, using Qwen3-8B. For the `w/ CoT mix' configuration, a CoT ratio of 0.3 is applied. All reported values are percentages.}
\label{table:seq-level-results-8b-math}
\small
\setlength{\tabcolsep}{7pt}
\renewcommand{\arraystretch}{1.1}

\begin{tabular}{ll *{8}{c}}
\toprule
 & & \multicolumn{2}{c}{\bf Static sampling} & \multicolumn{2}{c}{\bf Sequence-level adapter} \\
\cmidrule(lr){3-4}\cmidrule(lr){5-6}
\multicolumn{1}{c}{\bf Metric} & \multicolumn{1}{c}{\bf Setting}
& \multicolumn{1}{c}{ Best }
& \multicolumn{1}{c}{\makecell{Mixed}}
& \multicolumn{1}{c}{\makecell{w/o budget}}
& \multicolumn{1}{c}{\makecell{w/ budget}} \\
\midrule
\multirow{2}{*}{Pass@1}
 & w/o CoT &  72.86 \(\pm\) 1.23 & 72.38 \(\pm\) 1.24 & 73.20 \(\pm\) 1.23  & \bestcell{73.67 \(\pm\) 1.22 } \\
 & & &  \\
 & mix CoT &  73.02 \(\pm\) 1.23 & 72.92 \(\pm\) 1.23 & 73.80 \(\pm\) 1.22  & \bestcell{74.07 \(\pm\) 1.21} \\
 & \\
\addlinespace
\multirow{2}{*}{\makecell{Pass@8}}
 & w/o CoT &   78.38 \(\pm\)1.14  &  77.10 \(\pm\)1.16  &   78.20 \(\pm\) 1.14  &  \bestcell{78.80 \(\pm\) 1.13} \\
 & mix CoT &   78.98 \(\pm\) 1.13 &  78.58 \(\pm\) 1.14 & 79.00 \(\pm\) 1.13 &  \bestcell{ \textbf{79.13} \(\pm\) 1.13} \\
 \addlinespace
\bottomrule
\end{tabular}
\end{table*}

\begin{table*}[ht]
\centering
\caption{\textbf{Comparison of sampling strategies on CodeContests.}: vs. the \textbf{sequence}-level sampling adapter under two inference settings (without/with a mix of CoT prompts), using Qwen3-8B. For `w/ CoT mix', we use a CoT ratio of 0.3. }
\label{table:seq-level-results-8b-code}
\small
\setlength{\tabcolsep}{7pt}
\renewcommand{\arraystretch}{1.1}

\begin{tabular}{ll *{8}{c}}
\toprule
 & & \multicolumn{2}{c}{\bf Static sampling} & \multicolumn{2}{c}{\bf Sequence-level adapter} \\
\cmidrule(lr){3-4}\cmidrule(lr){5-6}
\multicolumn{1}{c}{\bf Metric} & \multicolumn{1}{c}{\bf Setting}
& \multicolumn{1}{c}{ Best }
& \multicolumn{1}{c}{\makecell{Mixed}}
& \multicolumn{1}{c}{\makecell{w/o budget}}
& \multicolumn{1}{c}{\makecell{w/ budget}} \\
\midrule
\multirow{2}{*}{Pass@1}
 & w/o CoT &  12.80 \(\pm\) 2.39 & 12.00 \(\pm\) 2.33 & \bestcell{ 18.80 \(\pm\) 2.80 } &  17.90 \(\pm\) 2.74 \\
 & mix CoT &  19.70 \(\pm\) 2.85 &  14.50 \(\pm\) 2.52  & 22.20 \(\pm\) 2.97  &  \bestcell{23.10 \(\pm\) 3.02} \\
\addlinespace
\multirow{2}{*}{\makecell{Pass@8}}
 & w/o CoT  &  23.10 \(\pm\) 3.02 &  22.23 \(\pm\)2.98 & 26.50 \(\pm\) 3.16 & \bestcell{27.40  \(\pm\) 3.19 } \\
 & mix CoT  &  29.10 \(\pm\) 3.25 &  23.90 \(\pm\) 3.05 & 32.50 \(\pm\) 3.35  &  \bestcell{\textbf{34.20} \(\pm\) 3.40 } \\
 \addlinespace
\bottomrule
\end{tabular}
\end{table*}

\begin{table}[H]
\centering
\caption{\textbf{Comparison of sampling strategies on MATH.} the \textbf{sequence}-level sampling adapter under two inference settings (without/with a mix of CoT prompts). 0.3.  }
\label{table:seq-level-results-1.5b-math}
\small
\begin{tabular}{ll *{8}{c}}
\toprule
 & & \multicolumn{2}{c}{\bf Static sampling} & \multicolumn{2}{c}{\bf Sequence-level adapter} \\
\cmidrule(lr){3-4}\cmidrule(lr){5-6}
\multicolumn{1}{c}{\bf Metric} & \multicolumn{1}{c}{\bf Setting}
& \multicolumn{1}{c}{Best}
& \multicolumn{1}{c}{\makecell{Mixed}}
& \multicolumn{1}{c}{\makecell{Just\\4 actions}}
& \multicolumn{1}{c}{\makecell{w/ mix of\\inference strategy}} \\
\midrule
\multirow{2}{*}{Pass@1}
 & w/o CoT & 61.52 \(\pm\) 1.35 & 56.42 \(\pm\) 1.37 & 61.52 \(\pm\) 1.35 & \bestcell{62.62 \(\pm\) 1.34 }\\
 & mix CoT & 62.10 \(\pm\) 1.34 & 59.00 \(\pm\) 1.36 & 62.76 \(\pm\) 1.34  & \bestcell{\textbf{63.20} \(\pm\) 1.34} \\
\addlinespace
\multirow{2}{*}{\makecell{Pass@8}}
 & w/o CoT & 61.52 \(\pm\) 1.35 & 66.74 \(\pm\) 1.31 & 68.25 \(\pm\) 1.29 & \bestcell{69.66 \(\pm\) 1.27} \\
 & mix  CoT  & 63.00 \(\pm\) 1.34 & 67.10 \(\pm\) 1.30 & 69.32 \(\pm\) 1.28 & \bestcell{\textbf{70.88} \(\pm\) 1.26} \\
 \addlinespace
\bottomrule
\end{tabular}
\end{table}

\subsection{Generalization}
In \Cref{table:seq-gen-to-coding}, we present the evaluation results on CodeContests of a sequence adapter trained on MATH to validate the out-of-domain generalization ability. 
\begin{table*}[h]
\centering
\caption{Coding: performance comparison of \textbf{sequence}-level adapter trained on MATH train versus baseline numbers.}
\label{table:seq-gen-to-coding}
\small
\begin{tabular}{ll *{4}{c}}
\toprule
 & & \multicolumn{2}{c}{\bf Static sampling} & \multicolumn{2}{c}{\bf Adapter} \\
\cmidrule(lr){3-4}\cmidrule(lr){5-6}
\multicolumn{1}{c}{\bf Metric} & \multicolumn{1}{c}{\bf Setting}
& \multicolumn{1}{c}{Best }
& \multicolumn{1}{c}{Mixed } 
& \multicolumn{1}{c}{w/o budget} 
\\
\midrule
\multirow{2}{*}{Pass@1} & w/o CoT & 11.43 \(\pm\) 2.28 & 10.53 \(\pm\) 2.20 & \bestcell{12.80 \(\pm\) 2.39 } &   \\
 & mix CoT  & 13.97 \(\pm\) 2.48 & 14.80 \(\pm\) 2.54 & \bestcell{17.90 \(\pm\) 2.74 } &   \\
\bottomrule
\end{tabular}
\end{table*}



\subsection{Qualitative Results}
\label{app:qual}

This appendix provides qualitative evidence about how the learned adapter behaves at the token level.
Our goal here is \emph{not} to reverse-engineer a complete, human-interpretable rule for the policy, but to
(1) sanity-check that the adapter meaningfully changes token-level stochasticity relative to static decoding,
and (2) probe whether a small set of simple heuristics (e.g., entropy-thresholding) could plausibly explain the gains.
Overall, we observe consistent trends (e.g., differential treatment of low- vs.\ high-entropy tokens), but we do not
find a single low-dimensional qualitative pattern that fully characterizes the learned behavior.

\begin{figure}[ht]
    \centering
    \includegraphics[width=.5\textwidth]{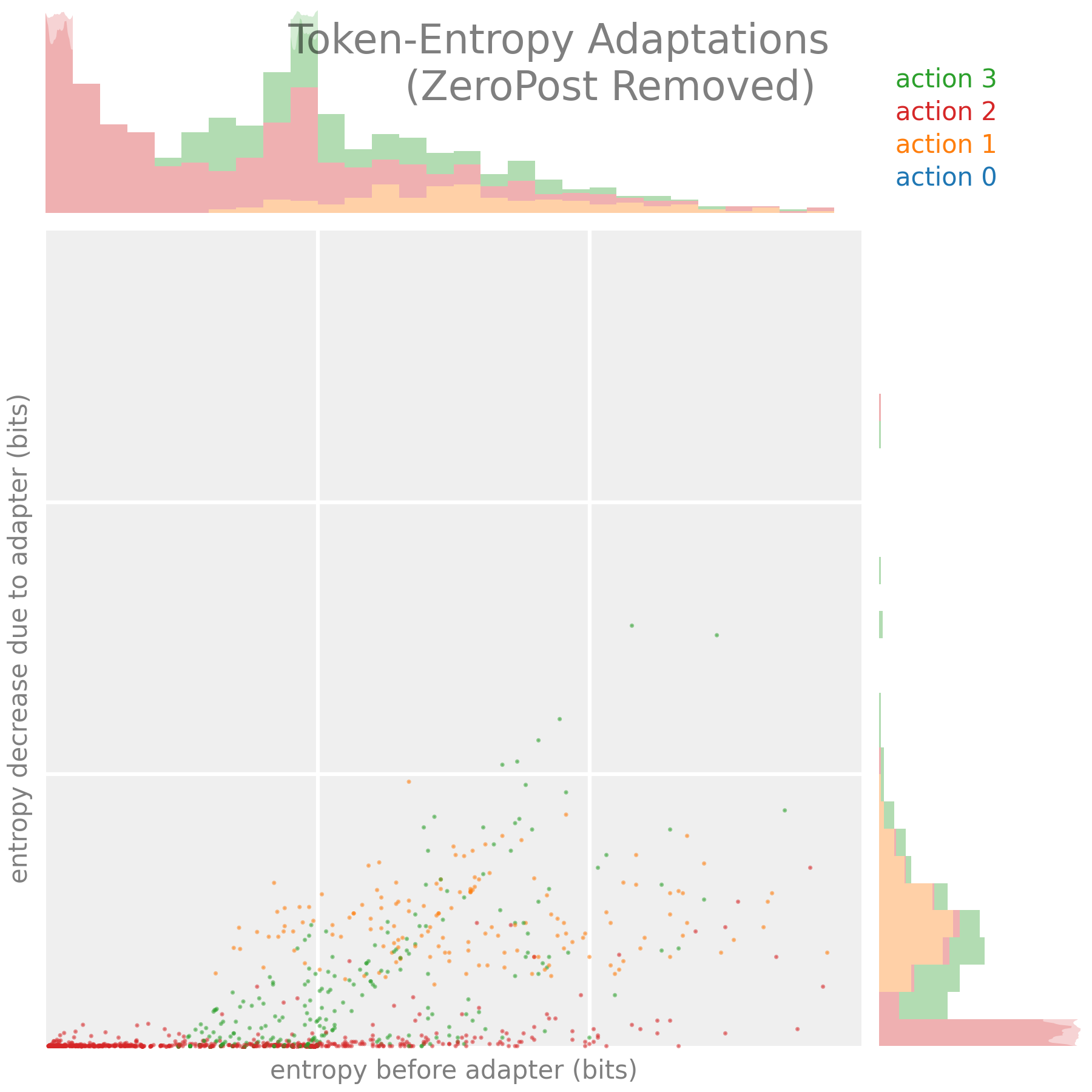}
    \caption{\textbf{Entropy modulation under token-level control.} Token entropy distributions for a representative static strategy
    versus the learned token-level adapter, using the temperature-only action set from \Cref{tab:math-token}. Action 0 is greedy; larger
    action indices correspond to higher temperature. Qualitatively, the adapter more often preserves stochasticity on higher-entropy tokens,
    while collapsing many low-entropy tokens to near-deterministic behavior.}
    \label{fig:tok-entropy}
\end{figure}

\paragraph{Entropy modulation.}
Figure~\ref{fig:tok-entropy} compares token entropy distributions under static decoding and the learned token-level adapter.
A consistent qualitative trend is that the adapter collapses \emph{low-entropy} tokens to near-deterministic behavior more often,
while \emph{high-entropy} tokens retain stochasticity more frequently. Importantly, this trend should be interpreted as
a correlational signature of the learned policy rather than a complete explanation of its success: in the main paper we show
that an entropy-only observation is insufficient to match the full adapter's performance.

\begin{figure*}[t]
\centering
\begin{subfigure}[b]{0.4\textwidth}
    \centering
    \includegraphics[width=\textwidth]{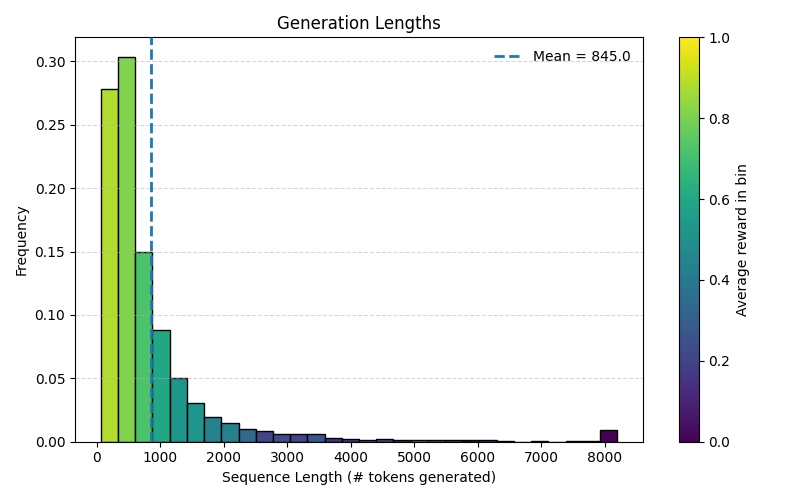}
    \caption{Static decoding (representative strategy).}
    \label{fig:length-static}
\end{subfigure}
\hfill
\begin{subfigure}[b]{0.4\textwidth}
    \centering
    \includegraphics[width=\textwidth]{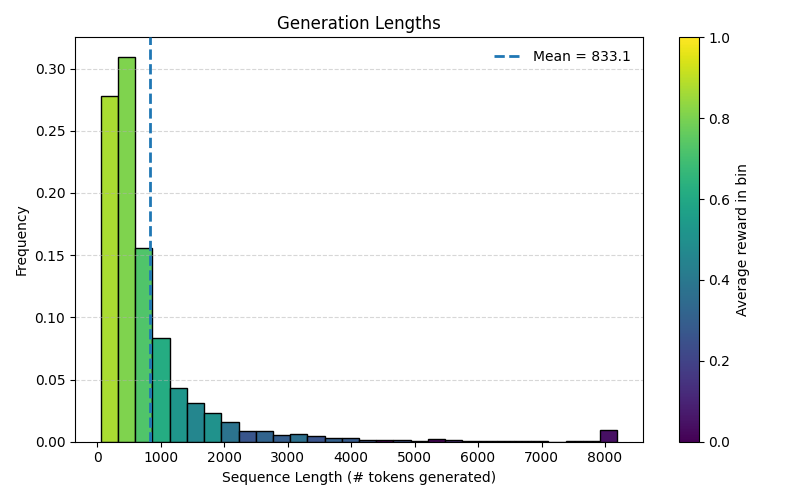}
    \caption{Sequence-level adapter.}
    \label{fig:length-seq}
\end{subfigure}
\hfill
\caption{\textbf{Generation length distributions.} Generation length statistics for a representative static strategy and the learned adapters.
Length shifts are present but not extreme, suggesting the token-level gains are not explained solely by systematic truncation/verbosity changes.}
\label{fig:length-distribution}
\end{figure*}

\paragraph{Generation length statistics.}
Figure~\ref{fig:length-distribution} reports generation-length distributions for a representative static strategy and for the learned
sequence- and token-level adapters. We include this as a coarse behavioral check: large performance gains can sometimes be confounded by
systematic changes in generation length (e.g., consistently shorter outputs). Here, lengths shift only moderately, suggesting the gains are
not solely a trivial artifact of length.

\paragraph{Example generation (action traces).}
To make token-level control concrete, we show an example generation with tokens colored by the \emph{current} selected action
(\Cref{text:example-now}) and the \emph{next} action (\Cref{text:example-next}). Black indicates tokens with very low entropy,
where the choice of decoding action rarely changes the sampled token; other colors correspond to the four actions in \Cref{tab:math-token}.
While these examples illustrate that the policy switches actions within a single trajectory, we did not find a simple, global rule
(e.g., a single entropy threshold) that explains when switches occur across prompts.

\begin{tcolorbox}[title=Example Generation (Current Action), breakable, fontupper=\small]
\label{text:example-now}
{\color{darkgreen} To}
{\color{lightblue} determine}
{\color{lightblue} how}
{\color{black} many}
{\color{black} seconds}
{\color{black} it}
{\color{black} takes}
{\color{black} for}
{\color{black} the}
{\color{darkgreen} plane}
{\color{black} to}
{\color{black} reach}
{\color{black} an}
{\color{black} altitude}
{\color{black} of}
{\color{black} }{\color{black}1}{\color{black}2}{\color{black},}{\color{black}0}{\color{black}0}{\color{black}0}
{\color{black} feet}{\color{lightblue},}
{\color{black} we}
{\color{orange} need}
{\color{black} to}
{\color{lightblue} analyze}
{\color{black} the}
{\color{darkgreen} pattern}
{\color{black} of}
{\color{lightblue} its}
{\color{lightblue} climbing}{\color{red}.}
{\color{black} The}
{\color{darkgreen} plane}
{\color{black} climbs}
{\color{black} }{\color{black}1}{\color{black}0}{\color{black}0}
{\color{black} feet}
{\color{orange} in}
{\color{black} the}
{\color{black} first}
{\color{black} second}{\color{black},}
{\color{black} }{\color{black}2}{\color{black}0}{\color{black}0}
{\color{black} feet}
{\color{black} in}
{\color{black} the}
{\color{black} second}
{\color{black} second}{\color{black},}
{\color{black} }{\color{black}3}{\color{black}0}{\color{black}0}
{\color{black} feet}
{\color{black} in}
{\color{black} the}
{\color{black} third}
{\color{black} second}{\color{black},}
{\color{black} and}
{\color{black} so}
{\color{black} on}{\color{black}.}
{\color{black} This}
{\color{orange} forms}
{\color{black} an}
{\color{black} arithmetic}
{\color{red} series}
{\color{black} where}
{\color{black} the}
{\color{black} first}
{\color{black} term}
{\color{black} \textbackslash(}{\color{black}a}{\color{darkgreen}\_}{\color{black}1}{\color{red}\textbackslash)}
{\color{black} is}
{\color{black} }{\color{black}1}{\color{black}0}{\color{black}0}
{\color{lightblue} feet}
{\color{orange} and}
{\color{black} the}
{\color{black} common}
{\color{black} difference}
{\color{black} \textbackslash(}{\color{black}d}{\color{black}\textbackslash)}
{\color{black} is}
{\color{orange} also}
{\color{black} }{\color{black}1}{\color{black}0}{\color{black}0}
{\color{black} feet}{\color{black}.

}
\vspace{\baselineskip}
{\color{black}The}
{\color{lightblue} total}
{\color{darkgreen} altitude}
{\color{black} climbed}
{\color{black} after}
{\color{black} \textbackslash(}{\color{black}n}{\color{black}\textbackslash)}
{\color{black} seconds}
{\color{red} is}
{\color{black} the}
{\color{black} sum}
{\color{black} of}
{\color{black} the}
{\color{black} first}{\color{black} \textbackslash(}{\color{black}n}{\color{black}\textbackslash)}{\color{black} terms}
{\color{black} of}
{\color{black} this}
{\color{black} arithmetic}
{\color{black} series}{\color{black}.}
{\color{black} The}
{\color{red} sum}
{\color{black} \textbackslash(}{\color{black}S}{\color{black}\_n}{\color{black}\textbackslash)}{\color{black} of}{\color{black} the}{\color{black} first}{\color{black} \textbackslash(}{\color{black}n}{\color{black}\textbackslash)}{\color{black} terms}
{\color{black} of}
{\color{black} an}
{\color{black} arithmetic}
{\color{black} series}
{\color{black} is}
{\color{black} given}
{\color{black} by}
{\color{darkgreen} the}
{\color{black} formula}{\color{black}:
}{\color{red}\textbackslash[}{\color{black} S}{\color{black}\_n}{\color{black} =}{\color{black} \textbackslash}{\color{black}frac}{\color{black}\{n}{\color{black}\}\{}{\color{black}2}{\color{black}\}}{\color{darkgreen} (}{\color{black}2}{\color{black}a}{\color{black}\_}{\color{black}1}{\color{black} +}{\color{black} (}{\color{black}n}{\color{black}-}{\color{black}1}{\color{black})d}{\color{black})}{\color{black} \textbackslash}{\color{orange}]

}

\vspace{\baselineskip}
{\color{orange}Sub}{\color{black}stit}{\color{black}uting}
{\color{darkgreen} the}
{\color{orange} given}
{\color{black} values}{\color{black} \textbackslash(}{\color{black}a}{\color{black}\_}{\color{black}1}{\color{black} =}{\color{black} }{\color{black}1}{\color{black}0}{\color{black}0}{\color{black}\textbackslash)}
{\color{darkgreen} and}
{\color{black} \textbackslash(}{\color{black}d}{\color{black} =}{\color{black} }{\color{black}1}{\color{black}0}{\color{black}0}{\color{lightblue}\textbackslash}{\color{black}),}
{\color{black} we}
{\color{black} get}{\color{black}:
}{\color{black}\textbackslash[}{\color{black} S}{\color{black}\_n}
{\color{black} =}
{\color{black} \textbackslash}{\color{black}frac}{\color{black}\{n}{\color{black}\}\{}{\color{black}2}{\color{black}\}}{\color{black} (}{\color{black}2}{\color{black} \textbackslash}{\color{black}cdot}{\color{black} }{\color{black}1}{\color{black}0}{\color{black}0}{\color{black} +}{\color{black} (}{\color{black}n}{\color{black}-}{\color{black}1}{\color{black})}{\color{black} \textbackslash}{\color{black}cdot}{\color{black} }{\color{black}1}{\color{black}0}{\color{black}0}{\color{black})}
{\color{lightblue} =}
{\color{black} \textbackslash}{\color{black}frac}{\color{black}\{n}{\color{black}\}\{}{\color{black}2}{\color{black}\}}{\color{black} (}{\color{black}2}{\color{black}0}{\color{black}0}{\color{black} +}{\color{black} }{\color{black}1}{\color{black}0}{\color{black}0}{\color{lightblue}(n}{\color{black}-}{\color{black}1}{\color{black}))}
{\color{black} =}
{\color{black} \textbackslash}{\color{black}frac}{\color{black}\{n}{\color{black}\}\{}{\color{black}2}{\color{black}\}}{\color{black} (}{\color{lightblue}1}{\color{black}0}{\color{black}0}{\color{black}n}{\color{black} +}{\color{black} }{\color{black}1}{\color{black}0}{\color{black}0}{\color{black})}{\color{black} =}{\color{black} }{\color{black}5}{\color{black}0}{\color{black}n}{\color{black}(n}{\color{darkgreen} +}{\color{black} }{\color{black}1}{\color{black})}{\color{black} \textbackslash}{\color{black}]

}
\vspace{\baselineskip}
{\color{black}We}
{\color{black} need}
{\color{black} to}
{\color{black} find}
{\color{red} \textbackslash(}{\color{black}n}{\color{black}\textbackslash)}
{\color{black} such}
{\color{black} that}{\color{lightblue} \textbackslash(}{\color{black}S}{\color{black}\_n}{\color{black} =}{\color{black} }{\color{black}1}{\color{black}2}{\color{black}0}{\color{black}0}{\color{black}0}{\color{black}\textbackslash}{\color{black}):
}{\color{black}\textbackslash[}{\color{black} }{\color{black}5}{\color{black}0}{\color{black}n}{\color{black}(n}{\color{black} +}{\color{black} }{\color{black}1}{\color{black})}{\color{black} =}{\color{black} }{\color{black}1}{\color{black}2}{\color{black}0}{\color{black}0}{\color{black}0}{\color{black} \textbackslash}{\color{black}]
}{\color{black}\textbackslash[}{\color{black} n}{\color{black}(n}{\color{black} +}{\color{black} }{\color{black}1}{\color{black})}{\color{black} =}{\color{darkgreen} \textbackslash}{\color{black}frac}{\color{black}\{}{\color{black}1}{\color{black}2}{\color{black}0}{\color{black}0}{\color{black}0}{\color{black}\}\{}{\color{black}5}{\color{black}0}{\color{black}\}}{\color{black} =}{\color{black} }{\color{black}2}{\color{black}4}{\color{black}0}{\color{black} \textbackslash}{\color{orange}]

}
\vspace{\baselineskip}
{\color{darkgreen}Now}{\color{lightblue},}
{\color{black} we}
{\color{black} need}
{\color{black} to}
{\color{lightblue} solve}
{\color{black} the}
{\color{black} quadratic}
{\color{black} equation}
{\color{lightblue} \textbackslash(}{\color{black}n}{\color{black} \textbackslash \text{\textasciicircum}}{\color{black}2}{\color{black} +}{\color{black} n}{\color{black} -}{\color{black} }{\color{black}2}{\color{black}4}{\color{black}0}
{\color{black} =}
{\color{black} }{\color{black}0}{\color{black}\textbackslash}{\color{black}).}
{\color{black} We}
{\color{black} can}
{\color{lightblue} use}
{\color{black} the}
{\color{black} quadratic}
{\color{black} formula}{\color{black} \textbackslash(}{\color{black}n}{\color{black} =}{\color{black} \textbackslash}{\color{black}frac}{\color{black}\{-}{\color{black}b}{\color{black} \textbackslash}{\color{black}pm}{\color{black} \textbackslash}{\color{black}sqrt}{\color{black}\{}{\color{black}b}{\color{black} \textbackslash \text{\textasciicircum}}{\color{black}2}{\color{black} -}{\color{black} }{\color{black}4}{\color{black}ac}{\color{black}\}\}}{\color{black}\{}{\color{black}2}{\color{black}a}{\color{black}\}\textbackslash}{\color{black}),}
{\color{black} where}
{\color{black} \textbackslash(}{\color{black}a}{\color{black} =}{\color{black} }{\color{black}1}{\color{black}\textbackslash}{\color{black}),}{\color{black} \textbackslash(}{\color{black}b}
{\color{black} =}
{\color{black} }{\color{black}1}{\color{black}\textbackslash}{\color{black}),}{\color{black} and}{\color{black} \textbackslash(}{\color{black}c}{\color{black} =}{\color{black} -}{\color{black}2}{\color{black}4}{\color{black}0}{\color{black}\textbackslash}{\color{black}):
}{\color{black}\textbackslash[}{\color{black} n}{\color{black} =}{\color{black} \textbackslash}{\color{black}frac}{\color{black}\{-}{\color{black}1}{\color{black} \textbackslash}{\color{black}pm}{\color{black} \textbackslash}{\color{black}sqrt}{\color{black}\{}{\color{black}1}{\color{orange} \textbackslash \text{\textasciicircum}}{\color{black}2}{\color{black} -}{\color{black} }{\color{black}4}{\color{black} \textbackslash}{\color{black}cdot}{\color{black} }{\color{black}1}{\color{black} \textbackslash}{\color{black}cdot}{\color{black} (-}{\color{black}2}{\color{black}4}{\color{black}0}{\color{black})\}\}}{\color{black}\{}{\color{black}2}{\color{black} \textbackslash}{\color{black}cdot}{\color{black} }{\color{black}1}{\color{black}\}}
{\color{black} =}
{\color{black} \textbackslash}{\color{black}frac}{\color{black}\{-}{\color{black}1}{\color{black} \textbackslash}{\color{black}pm}{\color{black} \textbackslash}{\color{black}sqrt}{\color{black}\{}{\color{black}1}{\color{black} +}{\color{black} }{\color{black}9}{\color{black}6}{\color{black}0}{\color{black}\}\}}{\color{black}\{}{\color{black}2}{\color{black}\}}
{\color{black} =}
{\color{black} \textbackslash}{\color{black}frac}{\color{black}\{-}{\color{black}1}{\color{black} \textbackslash}{\color{black}pm}{\color{black} \textbackslash}{\color{black}sqrt}{\color{black}\{}{\color{black}9}{\color{black}6}{\color{black}1}{\color{black}\}\}}{\color{black}\{}{\color{black}2}{\color{black}\}}
{\color{black} =}
{\color{black} \textbackslash}{\color{black}frac}{\color{black}\{-}{\color{black}1}{\color{black} \textbackslash}{\color{black}pm}{\color{black} }{\color{black}3}{\color{black}1}{\color{black}\}\{}{\color{black}2}{\color{black}\}}{\color{black} \textbackslash}{\color{black}]

}
\vspace{\baselineskip}
{\color{black}This}
{\color{black} gives}
{\color{black} us}
{\color{black} two}
{\color{lightblue} solutions}{\color{black}:
}

{\color{black}\textbackslash[}{\color{black} n}{\color{black} =}{\color{black} \textbackslash}{\color{black}frac}{\color{red}\{}{\color{black}3}{\color{black}0}{\color{black}\}\{}{\color{black}2}{\color{black}\}}{\color{black} =}{\color{black} }{\color{black}1}{\color{black}5}{\color{black} \textbackslash}{\color{lightblue}]
}

{\color{black}\textbackslash[}{\color{black} n}{\color{black} =}{\color{black} \textbackslash}{\color{black}frac}{\color{black}\{-}{\color{black}3}{\color{black}2}{\color{black}\}\{}{\color{black}2}{\color{black}\}}{\color{black} =}{\color{black} -}{\color{black}1}{\color{black}6}{\color{black} \textbackslash}{\color{black}]

}
\vspace{\baselineskip}
{\color{black}Since}
{\color{black} \textbackslash(}{\color{black}n}{\color{black}\textbackslash)}
{\color{black} must}
{\color{black} be}
{\color{black} a}
{\color{black} positive}
{\color{black} integer}{\color{black},}
{\color{black} we}
{\color{orange} discard}
{\color{black} the}
{\color{black} negative}
{\color{black} solution}{\color{black}.}
{\color{black} Therefore}{\color{black},}
{\color{lightblue} \textbackslash(}{\color{black}n}{\color{black} =}{\color{black} }{\color{black}1}{\color{black}5}{\color{black}\textbackslash}{\color{black}).

}
\vspace{\baselineskip}
{\color{orange}To}
{\color{red} verify}{\color{black},}
{\color{black} we}
{\color{red} calculate}
{\color{black} the}
{\color{lightblue} sum}
{\color{black} of}
{\color{black} the}
{\color{darkgreen} first}
{\color{black} }{\color{black}1}{\color{black}5}
{\color{black} terms}
{\color{darkgreen} of}
{\color{black} the}
{\color{red} series}{\color{black}:
}{\color{black}\textbackslash[}{\color{black} S}{\color{black}\_\{}{\color{black}1}{\color{black}5}{\color{black}\}}
{\color{black} =}
{\color{black} }{\color{black}5}{\color{black}0}{\color{black} \textbackslash}{\color{black}cdot}{\color{black} }{\color{black}1}{\color{black}5}{\color{black} \textbackslash}{\color{black}cdot}{\color{lightblue} }{\color{black}1}{\color{black}6}
{\color{black} =}
{\color{black} }{\color{black}1}{\color{black}2}{\color{black}0}{\color{black}0}{\color{black}0}{\color{black} \textbackslash}{\color{black}]

}
\vspace{\baselineskip}
{\color{darkgreen}Thus}{\color{black},}
{\color{orange} it}
{\color{black} takes}
{\color{lightblue} the}
{\color{black} plane}
{\color{black} \textbackslash}{\color{black}(\textbackslash}{\color{black}boxed}{\color{black}\{}{\color{black}1}{\color{black}5}{\color{black} \}}{\color{black}\textbackslash}
\end{tcolorbox}

\begin{tcolorbox}[title=Example Generation (Next Action), breakable, fontupper=\small]
\label{text:example-next}
{\color{lightblue} To}
{\color{lightblue} determine}
{\color{orange} how}
{\color{black} many}
{\color{black} seconds}
{\color{black} it}
{\color{black} takes}
{\color{black} for}
{\color{black} the}
{\color{darkgreen} plane}
{\color{black} to}
{\color{black} reach}
{\color{black} an}
{\color{black} altitude}
{\color{black} of}
{\color{black} }{\color{black}1}{\color{black}2}{\color{black},}{\color{black}0}{\color{black}0}{\color{black}0}
{\color{black} feet}{\color{lightblue},}
{\color{black} we}
{\color{red} need}
{\color{black} to}
{\color{lightblue} analyze}
{\color{black} the}
{\color{lightblue} pattern}
{\color{black} of}
{\color{lightblue} its}
{\color{red} climbing}{\color{lightblue}.}
{\color{black} The}
{\color{lightblue} plane}
{\color{black} climbs}
{\color{black} }{\color{black}1}{\color{black}0}{\color{black}0}
{\color{black} feet}
{\color{lightblue} in}
{\color{black} the}
{\color{black} first}
{\color{black} second}{\color{black},}
{\color{black} }{\color{black}2}{\color{black}0}{\color{black}0}
{\color{black} feet}
{\color{black} in}
{\color{black} the}
{\color{black} second}
{\color{black} second}{\color{black},}
{\color{black} }{\color{black}3}{\color{black}0}{\color{black}0}
{\color{black} feet}
{\color{black} in}
{\color{black} the}{\color{black} third}{\color{black} second}{\color{black},}
{\color{black} and}
{\color{black} so}
{\color{black} on}{\color{black}.}
{\color{black} This}
{\color{red} forms}
{\color{black} an}
{\color{black} arithmetic}
{\color{lightblue} series}
{\color{black} where}
{\color{black} the}
{\color{black} first}
{\color{black} term}
{\color{black} \textbackslash(}{\color{black}a}{\color{lightblue}\_}{\color{black}1}{\color{lightblue}\textbackslash)}
{\color{black} is}
{\color{black} }{\color{black}1}{\color{black}0}{\color{black}0}
{\color{orange} feet}
{\color{lightblue} and}
{\color{black} the}
{\color{black} common}
{\color{black} difference}
{\color{black} \textbackslash(}{\color{black}d}{\color{black}\textbackslash)}
{\color{black} is}
{\color{lightblue} also}
{\color{black} }{\color{black}1}{\color{black}0}{\color{black}0}
{\color{black} feet}{\color{black}.

}
\vspace{\baselineskip}
{\color{black}The}
{\color{darkgreen} total}
{\color{red} altitude}
{\color{black} climbed}
{\color{black} after}{\color{black} \textbackslash(}{\color{black}n}{\color{black}\textbackslash)}
{\color{black} seconds}
{\color{darkgreen} is}
{\color{black} the}
{\color{black} sum}
{\color{black} of}
{\color{black} the}
{\color{black} first}
{\color{black} \textbackslash(}{\color{black}n}{\color{black}\textbackslash)}
{\color{black} terms}
{\color{black} of}
{\color{black} this}
{\color{black} arithmetic}
{\color{black} series}{\color{black}.}
{\color{black} The}
{\color{darkgreen} sum}
{\color{black} \textbackslash(}{\color{black}S}{\color{black}\_n}{\color{black}\textbackslash)}
{\color{black} of}
{\color{black} the}
{\color{black} first}
{\color{black} \textbackslash(}{\color{black}n}{\color{black}\textbackslash)}
{\color{black} terms}
{\color{black} of}
{\color{black} an}
{\color{black} arithmetic}
{\color{black} series}
{\color{black} is}
{\color{black} given}
{\color{black} by}
{\color{darkgreen} the}
{\color{black} formula}{\color{black}:
}{\color{lightblue}\textbackslash[}{\color{black} S}{\color{black}\_n}{\color{black} =}{\color{black} \textbackslash}{\color{black}frac}{\color{black}\{n}{\color{black}\}\{}{\color{black}2}{\color{black}\}}{\color{red} (}{\color{black}2}{\color{black}a}{\color{black}\_}{\color{black}1}{\color{black} +}{\color{black} (}{\color{black}n}{\color{black}-}{\color{black}1}{\color{black})d}{\color{black})}{\color{black} \textbackslash}{\color{orange}]

}
\vspace{\baselineskip}
{\color{red}Sub}{\color{black}stit}{\color{black}uting}
{\color{orange} the}
{\color{orange} given}
{\color{black} values}
{\color{black} \textbackslash(}{\color{black}a}{\color{black}\_}{\color{black}1}{\color{black} =}{\color{black} }{\color{black}1}{\color{black}0}{\color{black}0}{\color{black}\textbackslash)}
{\color{darkgreen} and}
{\color{black} \textbackslash(}{\color{black}d}
{\color{black} =}
{\color{black} }{\color{black}1}{\color{black}0}{\color{black}0}{\color{red}\textbackslash}{\color{black}),}
{\color{black} we}
{\color{black} get}{\color{black}:
}{\color{black}\textbackslash[}{\color{black} S}{\color{black}\_n}
{\color{black} =}
{\color{black} \textbackslash}{\color{black}frac}{\color{black}\{n}{\color{black}\}\{}{\color{black}2}{\color{black}\}}{\color{black} (}{\color{black}2}{\color{black} \textbackslash}{\color{black}cdot}{\color{black} }{\color{black}1}{\color{black}0}{\color{black}0}{\color{black} +}{\color{black} (}{\color{black}n}{\color{black}-}{\color{black}1}{\color{black})}{\color{black} \textbackslash}{\color{black}cdot}{\color{black} }{\color{black}1}{\color{black}0}{\color{black}0}{\color{black})}
{\color{darkgreen} =}
{\color{black} \textbackslash}{\color{black}frac}{\color{black}\{n}{\color{black}\}\{}{\color{black}2}{\color{black}\}}{\color{black} (}{\color{black}2}{\color{black}0}{\color{black}0}{\color{black} +}{\color{black} }{\color{black}1}{\color{black}0}{\color{black}0}{\color{lightblue}(n}{\color{black}-}{\color{black}1}{\color{black}))}{\color{black} =}{\color{black} \textbackslash}{\color{black}frac}{\color{black}\{n}{\color{black}\}\{}{\color{black}2}{\color{black}\}}{\color{black} (}{\color{orange}1}{\color{black}0}{\color{black}0}{\color{black}n}{\color{black} +}{\color{black} }{\color{black}1}{\color{black}0}{\color{black}0}{\color{black})}
{\color{black} =}
{\color{black} }{\color{black}5}{\color{black}0}{\color{black}n}{\color{black}(n}{\color{darkgreen} +}{\color{black} }{\color{black}1}{\color{black})}{\color{black} \textbackslash}{\color{black}]

}
\vspace{\baselineskip}
{\color{black}We}
{\color{black} need}
{\color{black} to}
{\color{black} find}
{\color{darkgreen} \textbackslash(}{\color{black}n}{\color{black}\textbackslash)}
{\color{black} such}
{\color{black} that}
{\color{lightblue} \textbackslash(}{\color{black}S}{\color{black}\_n}
{\color{black} =}
{\color{black} }{\color{black}1}{\color{black}2}{\color{black}0}{\color{black}0}{\color{black}0}{\color{black}\textbackslash}{\color{black}):
}{\color{black}\textbackslash[}{\color{black} }{\color{black}5}{\color{black}0}{\color{black}n}{\color{black}(n}{\color{black} +}{\color{black} }{\color{black}1}{\color{black})}{\color{black} =}{\color{black} }{\color{black}1}{\color{black}2}{\color{black}0}{\color{black}0}{\color{black}0}{\color{black} \textbackslash}{\color{black}]
}{\color{black}\textbackslash[}{\color{black} n}{\color{black}(n}{\color{black} +}{\color{black} }{\color{black}1}{\color{black})}
{\color{black} =}
{\color{darkgreen} \textbackslash}{\color{black}frac}{\color{black}\{}{\color{black}1}{\color{black}2}{\color{black}0}{\color{black}0}{\color{black}0}{\color{black}\}\{}{\color{black}5}{\color{black}0}{\color{black}\}}
{\color{black} =}
{\color{black} }{\color{black}2}{\color{black}4}{\color{black}0}{\color{black} \textbackslash}{\color{darkgreen}]

}
\vspace{\baselineskip}
{\color{lightblue}Now}{\color{darkgreen},}
{\color{black} we}
{\color{black} need}
{\color{black} to}
{\color{lightblue} solve}
{\color{black} the}
{\color{black} quadratic}
{\color{black} equation}

{\color{lightblue} \textbackslash(}{\color{black}n}{\color{black}\textbackslash \text{\textasciicircum}}{\color{black}2}{\color{black} +}{\color{black} n}{\color{black} -}{\color{black} }{\color{black}2}{\color{black}4}{\color{black}0}
{\color{black} =}
{\color{black} }{\color{black}0}{\color{black}\textbackslash}{\color{black}).}

{\color{black} We}
{\color{black} can}
{\color{orange} use}
{\color{black} the}
{\color{black} quadratic}
{\color{black} formula}

{\color{black} \textbackslash(}{\color{black}n}
{\color{black} =}
{\color{black} \textbackslash}{\color{black}frac}{\color{black}\{-}{\color{black}b}{\color{black} \textbackslash}{\color{black}pm}{\color{black} \textbackslash}{\color{black}sqrt}{\color{black}\{}{\color{black}b}{\color{black}\textbackslash \text{\textasciicircum}}
{\color{black}2}{\color{black} -}{\color{black} }{\color{black}4}{\color{black}ac}{\color{black}\}\}}{\color{black}\{}{\color{black}2}{\color{black}a}{\color{black}\}\textbackslash}{\color{black}),}

{\color{black} where}
{\color{black} \textbackslash(}{\color{black}a}
{\color{black} =}
{\color{black} }{\color{black}1}{\color{black}\textbackslash}{\color{black}),}{\color{black} \textbackslash(}{\color{black}b}{\color{black} =}{\color{black} }{\color{black}1}{\color{black}\textbackslash}{\color{black}),}
{\color{black} and}{\color{black} \textbackslash(}{\color{black}c}{\color{black} =}{\color{black} -}{\color{black}2}{\color{black}4}{\color{black}0}{\color{black}\textbackslash}{\color{black}):
}{\color{black}\textbackslash[}{\color{black} n}
{\color{black} =}
{\color{black} \textbackslash}{\color{black}frac}{\color{black}\{-}{\color{black}1}{\color{black} \textbackslash}{\color{black}pm}{\color{black} \textbackslash}{\color{black}sqrt}{\color{black}\{}{\color{black}1}{\color{orange}\textbackslash \text{\textasciicircum}}{\color{black}2}{\color{black} -}{\color{black} }{\color{black}4}{\color{black} \textbackslash}{\color{black}cdot}{\color{black} }{\color{black}1}{\color{black} \textbackslash}{\color{black}cdot}{\color{black} (-}{\color{black}2}{\color{black}4}{\color{black}0}{\color{black})\}\}}{\color{black}\{}{\color{black}2}
{\color{black} \textbackslash}{\color{black}cdot}
{\color{black} }{\color{black}1}{\color{black}\}}
{\color{black} =}
{\color{black} \textbackslash}{\color{black}frac}{\color{black}\{-}{\color{black}1}{\color{black} \textbackslash}{\color{black}pm}{\color{black} \textbackslash}{\color{black}sqrt}{\color{black}\{}{\color{black}1}{\color{black} +}{\color{black} }{\color{black}9}{\color{black}6}{\color{black}0}{\color{black}\}\}}{\color{black}\{}{\color{black}2}{\color{black}\}}{\color{black} =}{\color{black} \textbackslash}{\color{black}frac}{\color{black}\{-}{\color{black}1}{\color{black} \textbackslash}{\color{black}pm}{\color{black} \textbackslash}{\color{black}sqrt}{\color{black}\{}{\color{black}9}{\color{black}6}{\color{black}1}{\color{black}\}\}}{\color{black}\{}{\color{black}2}{\color{black}\}}
{\color{black} =}
{\color{black} \textbackslash}{\color{black}frac}{\color{black}\{-}{\color{black}1}{\color{black} \textbackslash}{\color{black}pm}{\color{black} }{\color{black}3}{\color{black}1}{\color{black}\}\{}{\color{black}2}{\color{black}\}}{\color{black} \textbackslash}{\color{black}]

}
\vspace{\baselineskip}
{\color{black}This}
{\color{black} gives}
{\color{black} us}
{\color{black} two}
{\color{lightblue} solutions}{\color{black}:
}

{\color{black}\textbackslash[}{\color{black} n}{\color{black} =}{\color{black} \textbackslash}{\color{black}frac}{\color{lightblue}\{}{\color{black}3}{\color{black}0}{\color{black}\}\{}{\color{black}2}{\color{black}\}}{\color{black} =}{\color{black} }{\color{black}1}{\color{black}5}{\color{black} \textbackslash}{\color{lightblue}]
}

{\color{black}\textbackslash[}{\color{black} n}
{\color{black} =}
{\color{black} \textbackslash}{\color{black}frac}{\color{black}\{-}{\color{black}3}{\color{black}2}{\color{black}\}\{}{\color{black}2}{\color{black}\}}
{\color{black} =}
{\color{black} -}{\color{black}1}{\color{black}6}{\color{black} \textbackslash}{\color{black}]

}
\vspace{\baselineskip}
{\color{black}Since}{\color{black} \textbackslash(}{\color{black}n}{\color{black}\textbackslash)}
{\color{black} must}
{\color{black} be}
{\color{black} a}
{\color{black} positive}
{\color{black} integer}{\color{black},}
{\color{black} we}
{\color{orange} discard}
{\color{black} the}
{\color{black} negative}
{\color{black} solution}{\color{black}.}
{\color{black} Therefore}{\color{black},}
{\color{lightblue} \textbackslash(}{\color{black}n}{\color{black} =}{\color{black} }{\color{black}1}{\color{black}5}{\color{black}\textbackslash}{\color{black}).

}
\vspace{\baselineskip}
{\color{red}To}
{\color{lightblue} verify}{\color{black},}
{\color{black} we}
{\color{lightblue} calculate}
{\color{black} the}
{\color{lightblue} sum}
{\color{black} of}
{\color{black} the}
{\color{orange} first}
{\color{black} }{\color{black}1}{\color{black}5}
{\color{black} terms}
{\color{orange} of}
{\color{black} the}
{\color{darkgreen} series}{\color{black}:
}{\color{black}\textbackslash[}{\color{black} S}{\color{black}\_\{}{\color{black}1}{\color{black}5}{\color{black}\}}
{\color{black} =}
{\color{black} }{\color{black}5}{\color{black}0}
{\color{black} \textbackslash}{\color{black}cdot}{\color{black} }
{\color{black}1}{\color{black}5}
{\color{black} \textbackslash}{\color{black}cdot}
{\color{orange} }{\color{black}1}{\color{black}6}
{\color{black} =}
{\color{black} }{\color{black}1}{\color{black}2}{\color{black}0}{\color{black}0}{\color{black}0}{\color{black} \textbackslash}{\color{black}]

}
\vspace{\baselineskip}
{\color{red}Thus}{\color{black},}
{\color{lightblue} it}
{\color{black} takes}
{\color{red} the}
{\color{black} plane}
{\color{black} \textbackslash}{\color{black}(\textbackslash}{\color{black}boxed}{\color{black}\{}{\color{black}1}{\color{black}5}{\color{black}\}}{\color{black}\textbackslash}
\end{tcolorbox}

\begin{figure*}[t]
\begin{center}
\includegraphics[width=0.8\textwidth]{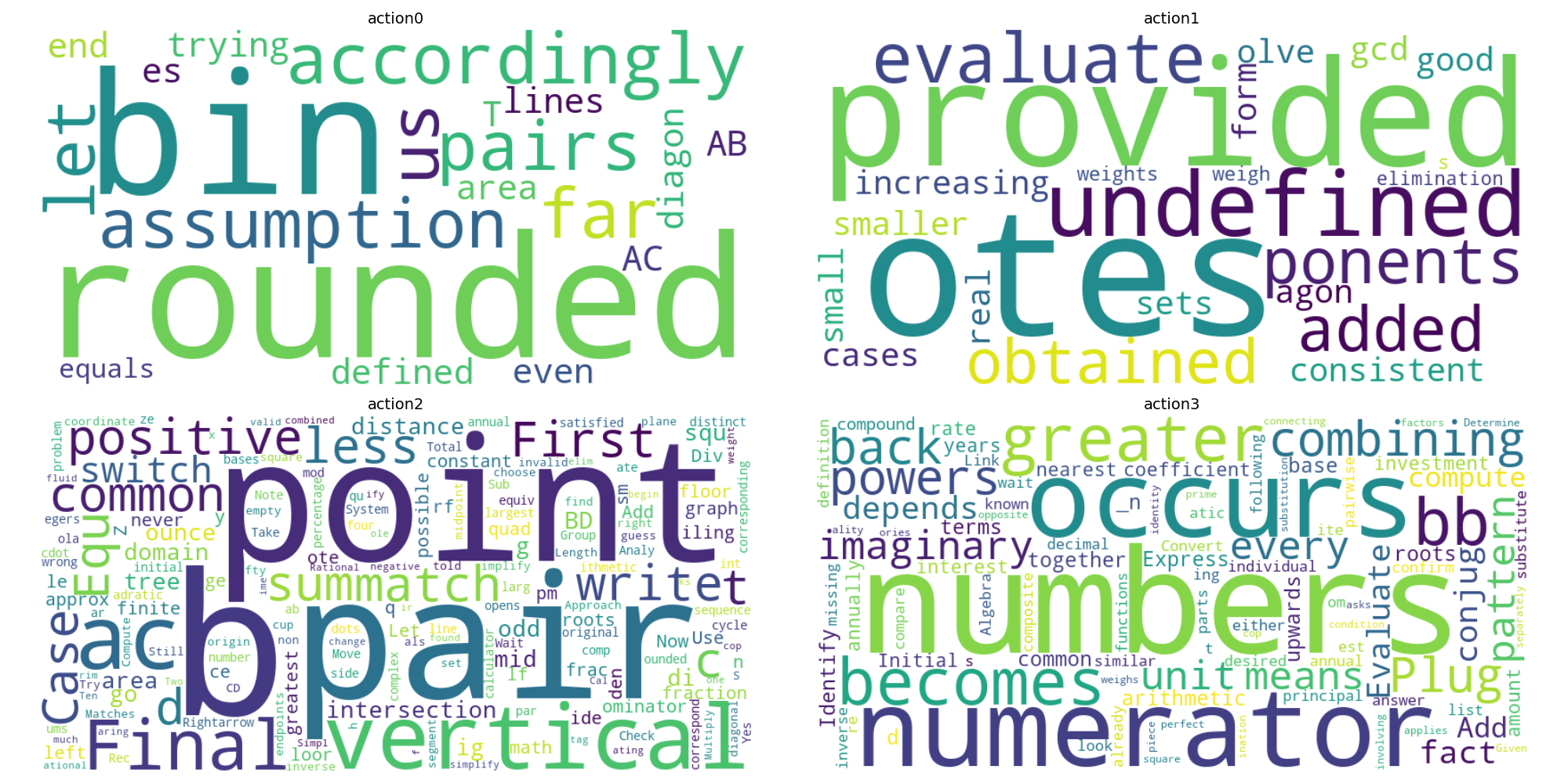}
\end{center}
\caption{\textbf{Word clouds by action.} Most frequent tokens under each temperature-based action from \Cref{tab:math-token}. This is a
lightweight diagnostic and does not by itself explain the policy's performance improvements.}
\label{fig:word-cloud}
\end{figure*}

\paragraph{Token content (word clouds).}
Figure~\ref{fig:word-cloud} visualizes frequent tokens under each action. We view this as an exploratory diagnostic rather than strong
evidence: word clouds can highlight obvious stylistic differences, but are insensitive to context and can be dominated by generic tokens.
In our runs, these visualizations did not reveal a clean, human-interpretable separation between actions beyond coarse stylistic effects.

\paragraph{Limitations.}
These qualitative analyses are intended to provide intuition and sanity checks, not a definitive mechanistic account of the learned policy.
In particular, the entropy trend in Fig.~\ref{fig:tok-entropy} is consistent with adaptive modulation of stochasticity, but the main paper's
entropy-only ablation indicates that richer contextual information is required to recover the full gains.

\section{Action Space Selection}
\label{appendix:action_selection}

This appendix describes the procedure used to construct a finite action space for the decoding adapters, along with empirical details of the selection process and the final strategies used in our experiments.

\paragraph{Candidate action pool.}
Let $\mathcal{S}$ denote a finite set of candidate decoding strategies. Each strategy $s \in \mathcal{S}$ corresponds to a fixed configuration of decoding hyperparameters (e.g., temperature, top-$k$, top-$p$, min-$p$). In our experiments, $\mathcal{S}$ is formed by the Cartesian product of the parameter ranges shown in Table~\ref{tab:sweep}, yielding a total of $5 \times 4 \times 3 \times 3 = 180$ candidate strategies. This pool is designed to span a broad range of decoding behaviors, from near-greedy to highly stochastic sampling.

\begin{table}[ht]
\centering
\caption{Sweep ranges of as the action space for our sampling adapter.}
\label{tab:sweep}
\small
\begin{tabular}{@{}ll@{}}
\toprule
\textbf{Parameter} & \textbf{Value} \\
\midrule
Temperature & $[0.3, 0.5, 0.75, 1.0, 1.25]$ \\
Top-$k$     & $[5, 10, 50, \text{``off"}]$ \\
Top-$p$     & $[0.9, 0.95, \text{``off"}]$ \\
Min-$p$     & $[0.1, 0.2, \text{``off"}]$ \\
\bottomrule
\end{tabular}
\end{table}

\paragraph{Reward model.}
Let $D = \{x_i\}_{i=1}^N$ denote a validation dataset. For each input $x_i$ and strategy $s$, we compute a verifiable terminal reward $r(x_i, s) \in \{0,1\}$ indicating task success (e.g., correctness of the generated solution). We define the empirical performance of a strategy as
\[
Q(s) = \frac{1}{N} \sum_{i=1}^N r(x_i, s).
\]

\paragraph{Objective.}
Our goal is not to identify a single best decoding strategy, but rather to select a small subset $S \subseteq \mathcal{S}$ such that, for each input, at least one strategy in $S$ performs well. Accordingly, we consider the following coverage-based objective:
\[
F(S) = \frac{1}{N} \sum_{i=1}^N \max_{s \in S} r(x_i, s),
\]
which corresponds to the expected performance of a best-of-$S$ decoder.

\paragraph{Greedy selection.}
Directly optimizing $F(S)$ is combinatorial. Following prior work on coverage-based selection~\cite{mavalankar2025aupair}, we construct $S$ using a greedy procedure: starting from $S_0 = \emptyset$, at each step we add
\[
s^\ast = \arg\max_{s \in \mathcal{S} \setminus S_k} \left( F(S_k \cup \{s\}) - F(S_k) \right),
\]
until a desired cardinality $|S| = K$ is reached. In practice, $K$ is chosen to balance expressivity of the action space against training stability of the downstream policy.

\begin{figure*}[h]
\begin{center}
\includegraphics[width=0.8\textwidth]{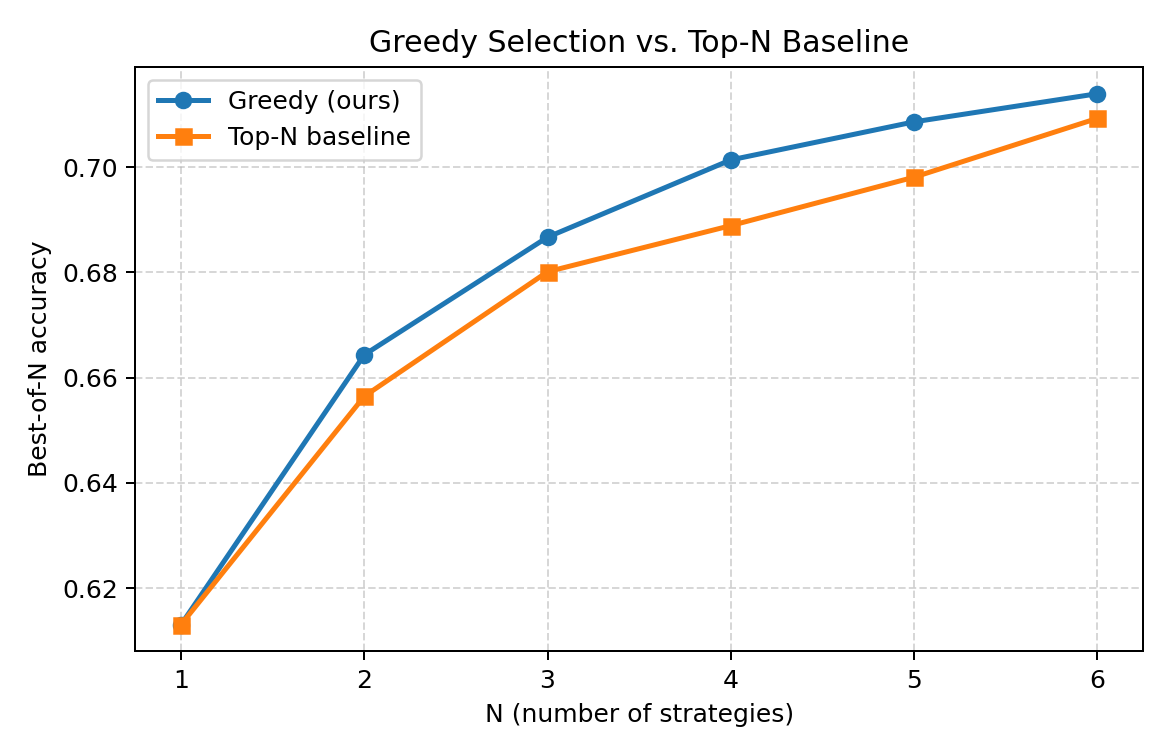}
\end{center}
\caption{Best of $N$ accuracy vs number of selected actions for greedy selection (our approach) vs the baseline of taking the top N actions. }
\label{fig:action_selection}
\end{figure*}

\paragraph{Empirical selection behavior.}
Figure~\ref{fig:action_selection} compares greedy coverage-based selection against a baseline that selects the top-$K$ strategies by average performance $Q(s)$. Greedy selection consistently achieves higher best-of-$K$ accuracy, indicating that it favors complementary strategies that succeed on different subsets of inputs rather than redundant high-average strategies. This behavior motivates the use of greedy coverage as a principled method for constructing a compact but expressive action space.

\paragraph{Selected strategies.}
Tables~\ref{tab:mix} and~\ref{tab:math-token} report the final sets of decoding strategies selected by the greedy procedure for the sequence-level and token-level experiments, respectively. At the sequence level, the selected strategies span a range of sampling behaviors and include both stochastic and near-greedy decoding. At the token level, the selected actions differ in temperature, reflecting the exploratory observation that stochasticity is the dominant axis along which complementary decoding behaviors emerge.

Below we provide the best set of strategies for math, coding, and the mixture of math and coding  evaluation sets. 

\begin{table}[h!]
\centering
\caption{Selected decoding strategies for \textbf{math} tasks.}
\begin{tabular}{cccccc}
\toprule
\textbf{Strategy} & \multicolumn{5}{c}{\textbf{Parameters}} \\
\cmidrule(lr){2-6}
 & \texttt{do\_sample} & \texttt{temperature} & \texttt{top\_k} & \texttt{top\_p} & \texttt{min\_p} \\
\midrule
1 & true  & 0.75 & --  & 0.9  & 0.1 \\
2 & true  & 1.0  & --  & 0.9  & --  \\
3 & true  & 1.25 & 10  & 0.9  & --  \\
4 & true  & 1.0  & 5   & 0.9  & --  \\
5 & true  & 1.25 & 50  & 0.9  & --  \\
6 & true  & 1.25 & --  & --   & --  \\
\bottomrule
\end{tabular}
\label{tab:math}
\end{table}

\begin{table}[h!]
\centering
\caption{Selected decoding strategies for \textbf{coding} tasks.}
\begin{tabular}{cccccc}
\toprule
\textbf{Strategy} & \multicolumn{5}{c}{\textbf{Parameters}} \\
\cmidrule(lr){2-6}
 & \texttt{do\_sample} & \texttt{temperature} & \texttt{top\_k} & \texttt{top\_p} & \texttt{min\_p} \\
\midrule
1 & true  & 1.25 & --  & 0.95 & 0.1 \\
2 & true  & 1.25 & --  & 0.95 & 0.2 \\
3 & true  & 1.0  & --  & 0.95 & 0.2 \\
4 & true  & 1.25 & 50  & 0.9  & 0.1 \\
5 & true  & 1.25 & 5   & 0.9  & --  \\
6 & true  & 0.75 & 5   & --   & 0.1 \\
\bottomrule
\end{tabular}
\label{tab:coding}
\end{table}

\begin{table}[ht]
\centering
\caption{Selected decoding strategies for a mix of \textbf{math and coding} tasks.}
\begin{tabular}{cccccc}
\toprule
\textbf{Strategy} & \multicolumn{5}{c}{\textbf{Parameters}} \\
\cmidrule(lr){2-6}
 & \texttt{do\_sample} & \texttt{temperature} & \texttt{top\_k} & \texttt{top\_p} & \texttt{min\_p} \\
\midrule
0 & true  & 0.5  & --   & --   & --   \\
1 & true  & 1.0  & --   & --   & --   \\
2 & true  & 1.25 & --   & --   & --   \\
3 & true  & 0.75 & 10   & --   & --   \\
4 & true  & 0.75 & 10   & 0.95 & 0.1  \\
5 & false & 0.0  & --   & --   & --   \\
\bottomrule
\end{tabular}
\label{tab:mix}
\end{table}

\begin{table}[ht]
\centering
\caption{Selected decoding strategies for token-level \textbf{math} tasks.}
\begin{tabular}{cccc}
\toprule
\textbf{Strategy} & \multicolumn{2}{c}{\textbf{Parameters}} \\
\cmidrule(lr){2-3}
 & \texttt{do\_sample} & \texttt{temperature}    \\
\midrule
0 & false  & 0     \\
1 & true  & 0.5       \\
2 & true  & 1.0      \\
3 & true  & 1.25     \\
\bottomrule
\end{tabular}
\label{tab:math-token}
\end{table}

\paragraph{Discussion.}
Although the candidate pool includes combinations of multiple sampling parameters, the greedy procedure tends to select strategies that differ primarily along a single axis of stochasticity. This observation motivates our use of temperature-based actions in the token-level experiments, while retaining a more general action space at the sequence level. We emphasize that this selection procedure is used only to define a finite action space prior to training; during policy learning, the adapter receives only task-level rewards and does not observe per-strategy performance estimates.

\section{Implementation Details}
\subsection{Hyperparameters}
In this section, we provide details on the implementation of our adapters.

\begin{table}[ht]
\centering
\caption{ Parameters of \textbf{token}-level sampling adapter.}
\label{table: tok}
\small
\begin{tabular}{ll}
\toprule
\multicolumn{1}{c}{\bf Parameter} & \multicolumn{1}{c}{\bf Value} \\
Architecture & MLP \\
Number of Layers & 3 \\
Activation &   SiLU \\
Dropout & 0.1 \\
Optimizer & Adam \\
format & bf16 \\
Learning rate scheduler & exponential \\
\bottomrule
\end{tabular}
\end{table}

We provide pseudocode for our sequence-level (\Cref{alg:seq_adapter}) and token-level (\Cref{alg:token_adapter}) Adapters below. 
For Sequence-Level Adapters, We use vLLM engine to fastly generate rollout. For token-Level, We use Ray Train~\cite{ray} and distributed data parallelism (DDP). We use past\_key\_values as an incremental KV-cache returned by the model each step, pass it back into the next forward for speed, and whenever some sequences finish early we “shrink” the cache to only the still-active batch rows via index\_select so it stays aligned with active\_idx.

\begin{algorithm}[t]
\caption{Sequence-Level Decoding Adapter}
\begin{algorithmic}[1]

\Function{SeqForward}{$\mathbf{e}, \mathbf{a}, T$}
  \If{$\text{a} \neq \varnothing$}
    \State $\text{z} \gets [\text{e} \parallel \mathrm{EmbedAux}(\text{a})]$
  \Else
    \State $\text{z} \gets \text{e}$
  \EndIf
  \State $\boldsymbol{\ell} \gets \mathrm{MLP}(\text{z})$ \Comment{logits over strategies}
  \State $\mathbf{p} \gets \mathrm{softmax}(\boldsymbol{\ell} / T)$
  \State \Return $\mathbf{p}$
\EndFunction

\Function{SeqSelectStrategy}{$\mathbf{e}, \mathbf{a}, T, \mathrm{deterministic}$}
  \State $\mathbf{p} \gets \Call{SeqForward}{\textbf{e}, \textbf{a}, T}$
  \If{\text{deterministic}}
    \State $s \gets \arg\max \mathbf{p}$
  \Else
    \State $s \sim \mathrm{Categorical}(\mathbf{p})$
  \EndIf
  \State \Return $s, \mathbf{p}$
\EndFunction

\end{algorithmic}
\label{alg:seq_adapter}
\end{algorithm}

\begin{algorithm}[t]
\caption{Token-Level Decoding Adapter}
\begin{algorithmic}[1]

\Function{TokForward}{$\mathbf{x}_t, T$}
  \State $\boldsymbol{\ell}_t \gets \mathrm{MLP}(\mathbf{x}_t)$ \Comment{logits over strategies}
  \State $\mathbf{p}_t \gets \mathrm{softmax}(\boldsymbol{\ell}_t / T)$
  \State \Return $\mathbf{p}_t$
\EndFunction

\Function{TokSelectStrategy}{$\mathbf{x}_t, T, \mathrm{deterministic}$}
  \State $\mathbf{p}_t \gets \Call{TokForward}{\mathbf{x}_t, T}$
  \If{\text{deterministic}}
    \State $s_t \gets \arg\max \mathbf{p}_t$
  \Else
    \State $s_t \sim \mathrm{Categorical}(\mathbf{p}_t)$
  \EndIf
  \State \Return $s_t, \mathbf{p}_t$
\EndFunction

\Function{TokFeatures}{$\mathrm{LLM}, \text{prefix}, t$}
  \State $\mathbf{h}_t \gets \mathrm{LastHidden}(\mathrm{LLM}(\text{prefix}))$
  \State \Return $\mathbf{x}_t \gets [\,\mathbf{h}_t \parallel \phi(t)\,]$ \Comment{optional aux/budget features}
\EndFunction

\Function{TokStep}{$\mathrm{LLM}, \text{prefix}, \text{transforms}, T, \mathrm{deterministic}$}
  \State $\mathbf{x}_t \gets \Call{TokFeatures}{\mathrm{LLM}, \text{prefix}, t}$
  \State $(s_t, \mathbf{p}_t) \gets \Call{TokSelectStrategy}{\mathbf{x}_t, T, \mathrm{deterministic}}$
  \State $y_t \gets \mathrm{SampleToken}(\mathrm{LLM}, \text{prefix}; \text{transforms}[s_t])$
  \State \Return $y_t, s_t, \mathbf{p}_t$
\EndFunction

\end{algorithmic}
\label{alg:token_adapter}
\end{algorithm}

\begin{figure*}[h]
\centering
\begin{subfigure}[b]{0.32\textwidth}
    \centering
    \includegraphics[width=\textwidth]{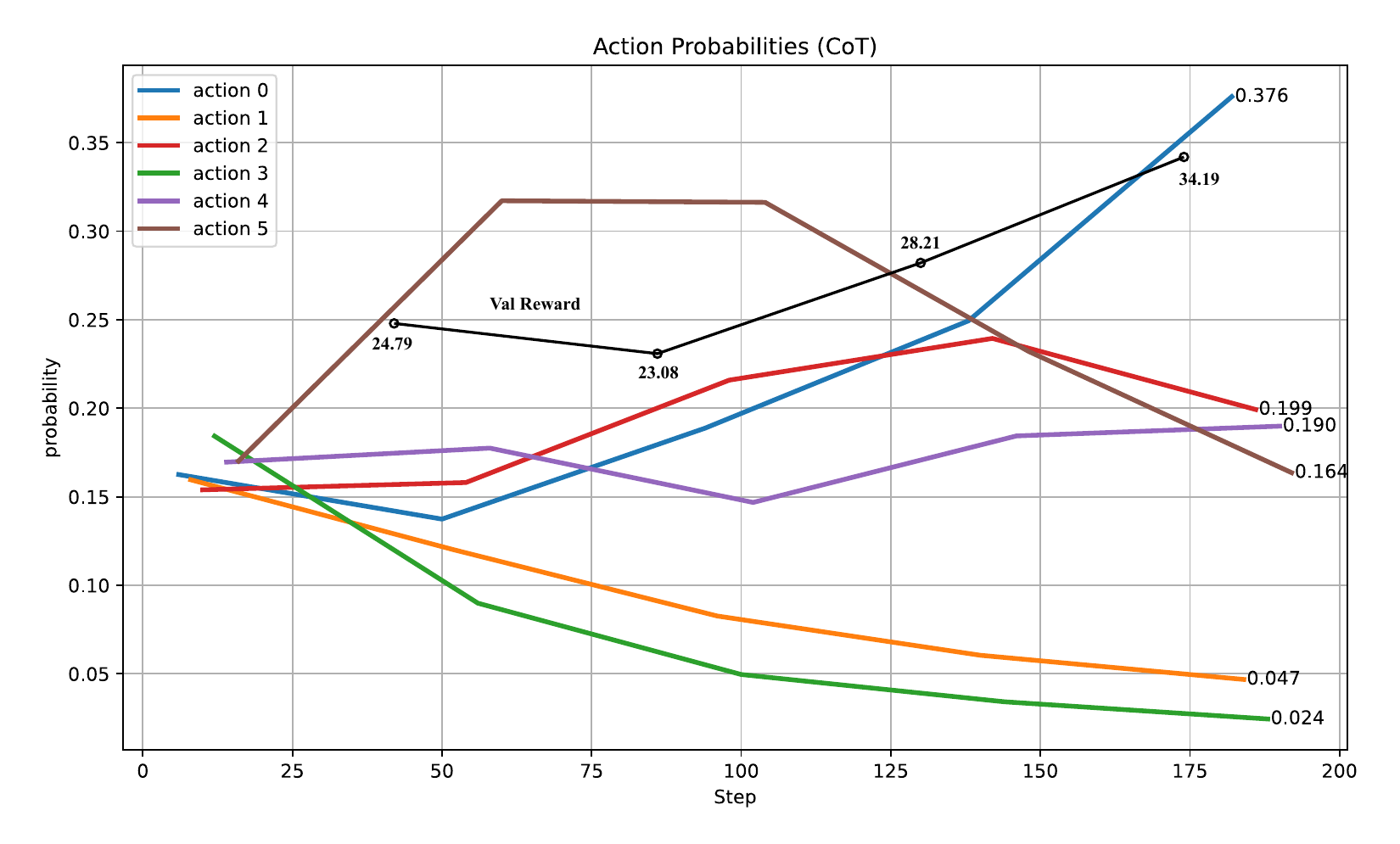}
    \caption{CoT action distribution}
    \label{fig:coding-cot}
\end{subfigure}
\hfill
\begin{subfigure}[b]{0.32\textwidth}
    \centering
    \includegraphics[width=\textwidth]{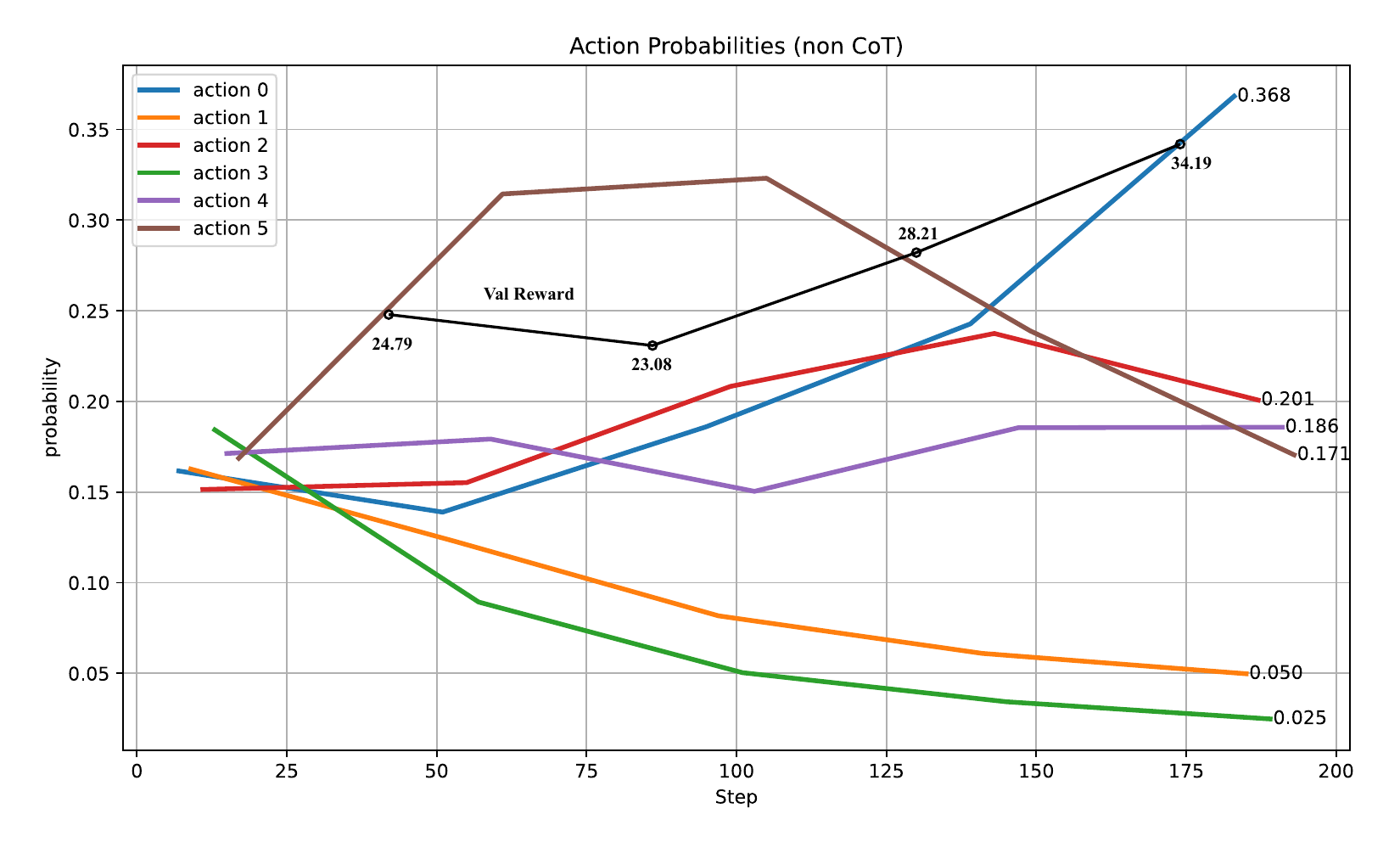}
    \caption{Non-CoT action distribution}
    \label{fig:coding-noncot}
\end{subfigure}
\hfill
\begin{subfigure}[b]{0.32\textwidth}
    \centering
    \includegraphics[width=\textwidth]{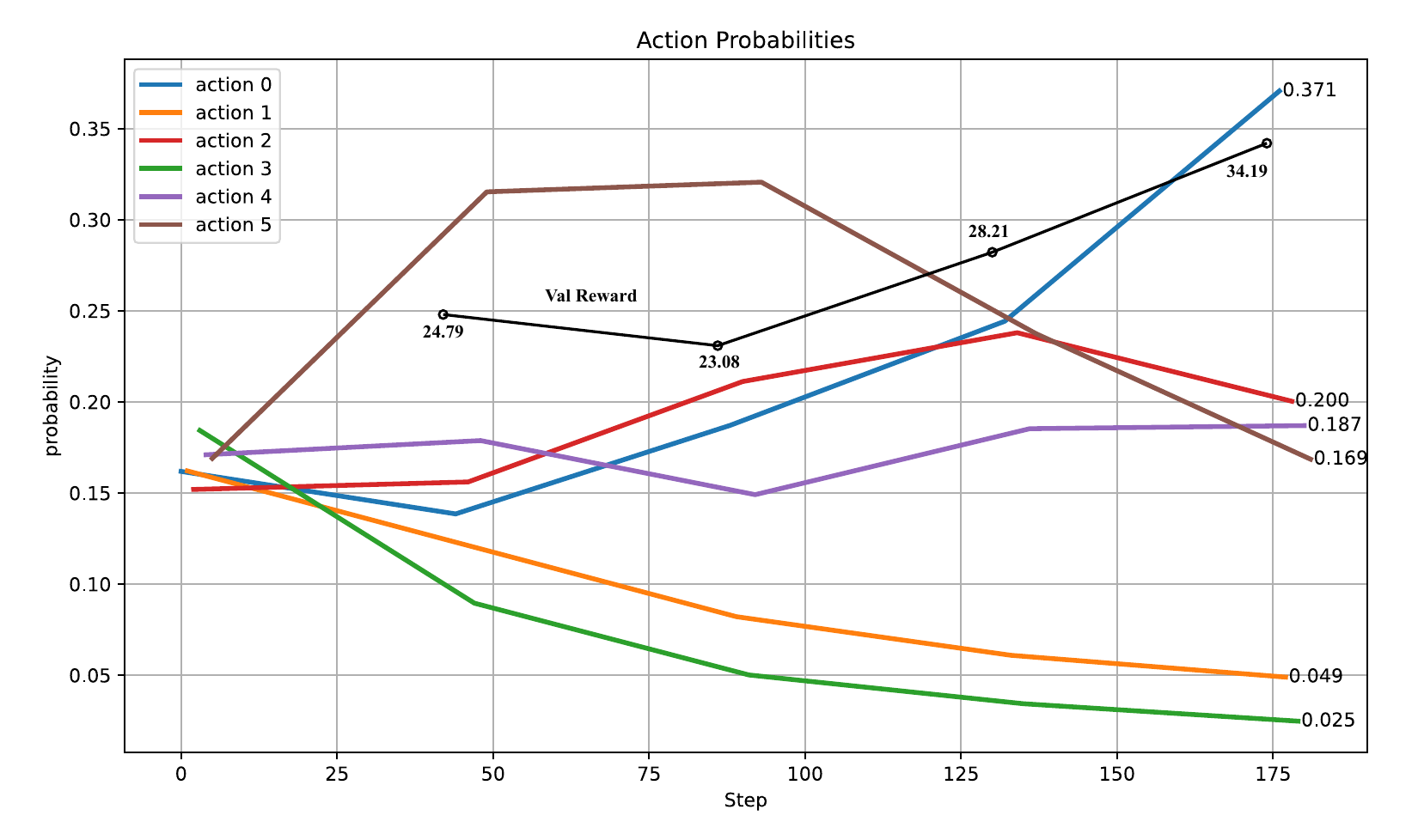}
    \caption{Overall action distribution}
    \label{fig:coding-all}
\end{subfigure}
\caption{ \textbf{Action distributions and val reward on CodeContests} for sampling strategies with sequence-level adapter. The actions are defined in \cref{tab:coding}. Axes are same as in \Cref{fig:math-action-distribution}. }
\label{fig:coding-action-distribution}
\end{figure*}

\begin{figure*}[ht]
\centering
\begin{subfigure}[b]{0.32\textwidth}
    \centering
    \includegraphics[width=\textwidth]{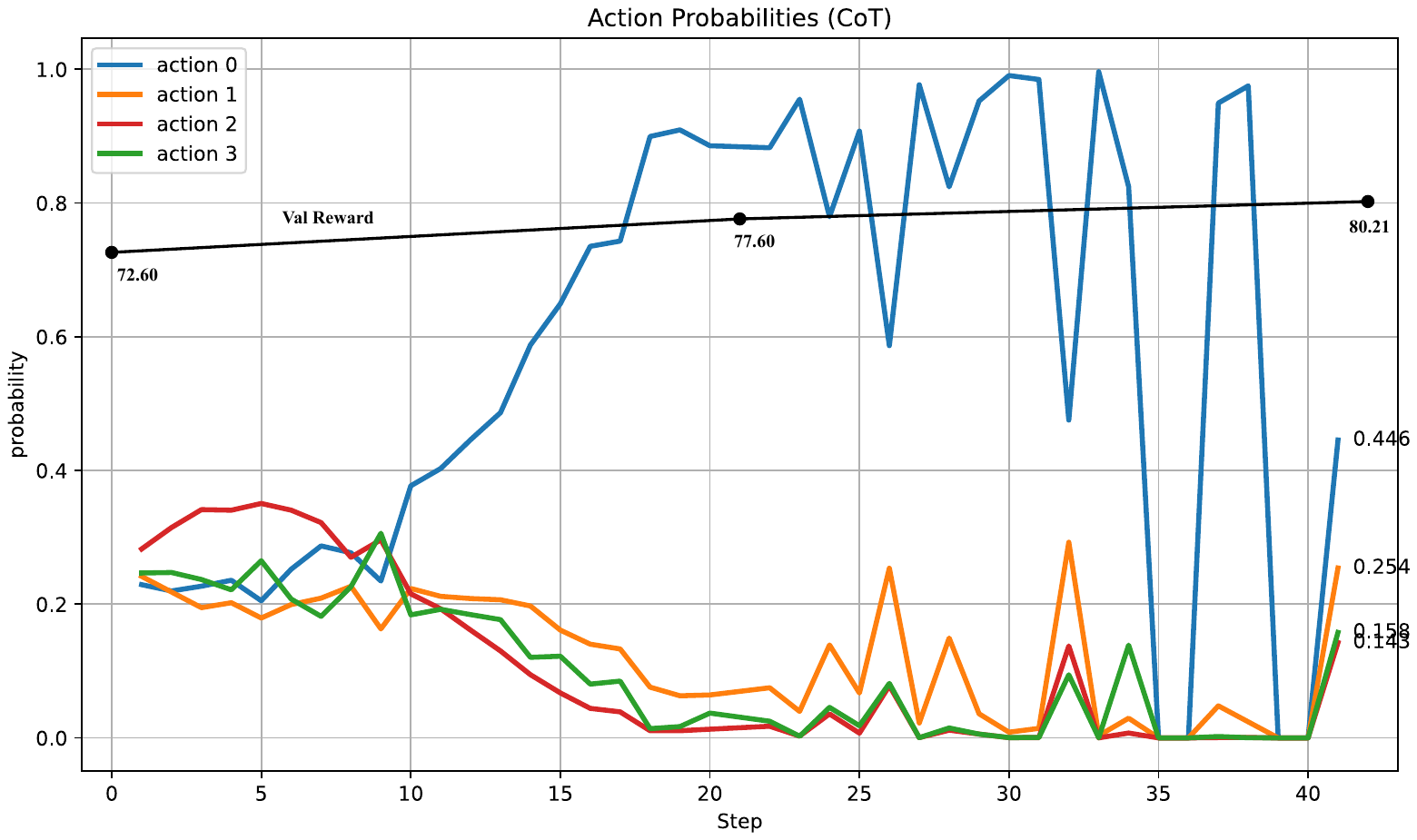}
    \caption{CoT action distribution}
    \label{fig:tok-cot}
\end{subfigure}
\hfill
\begin{subfigure}[b]{0.32\textwidth}
    \centering
    \includegraphics[width=\textwidth]{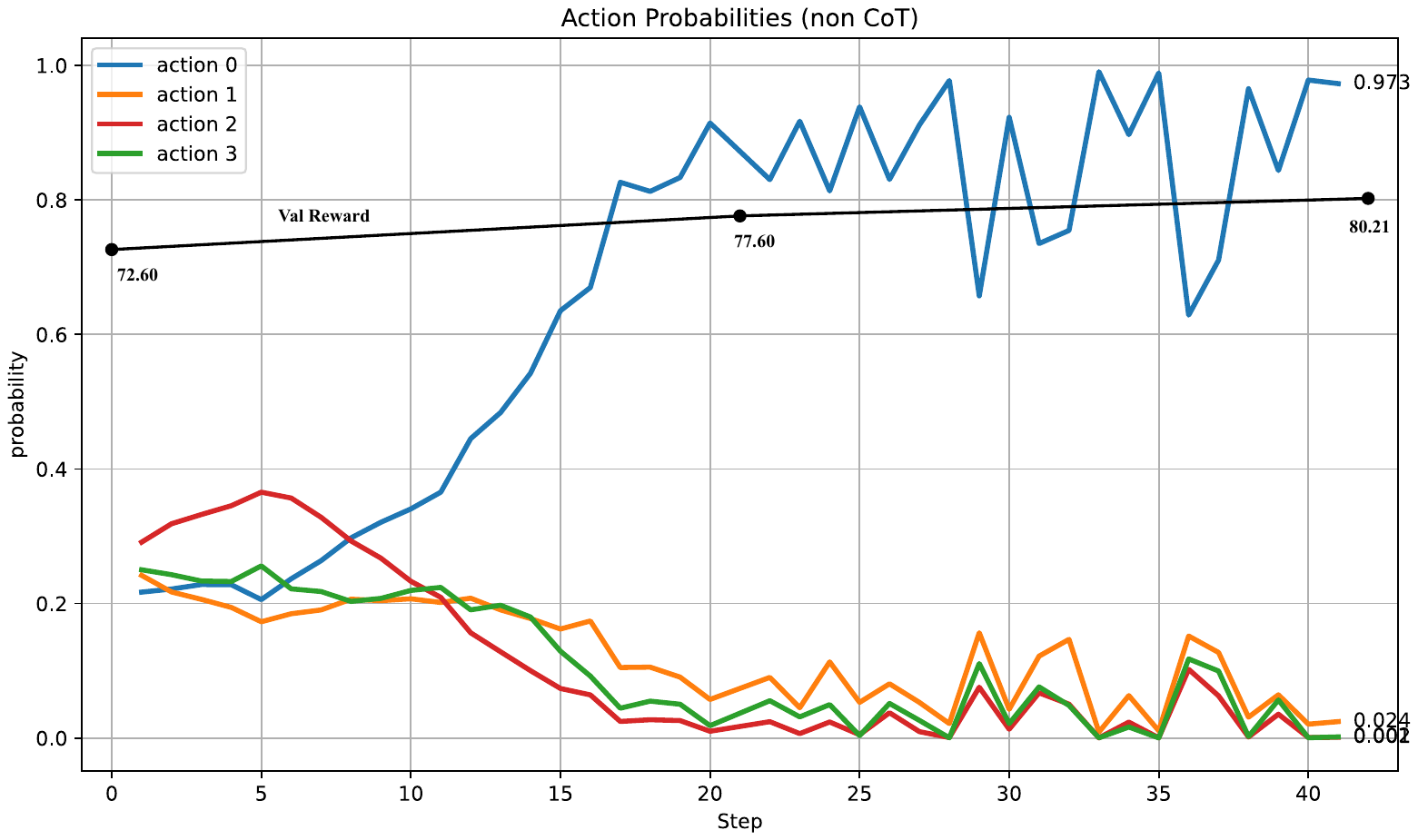}
    \caption{Non-CoT action distribution}
    \label{fig:tok-noncot}
\end{subfigure}
\hfill
\begin{subfigure}[b]{0.32\textwidth}
    \centering
    \includegraphics[width=\textwidth]{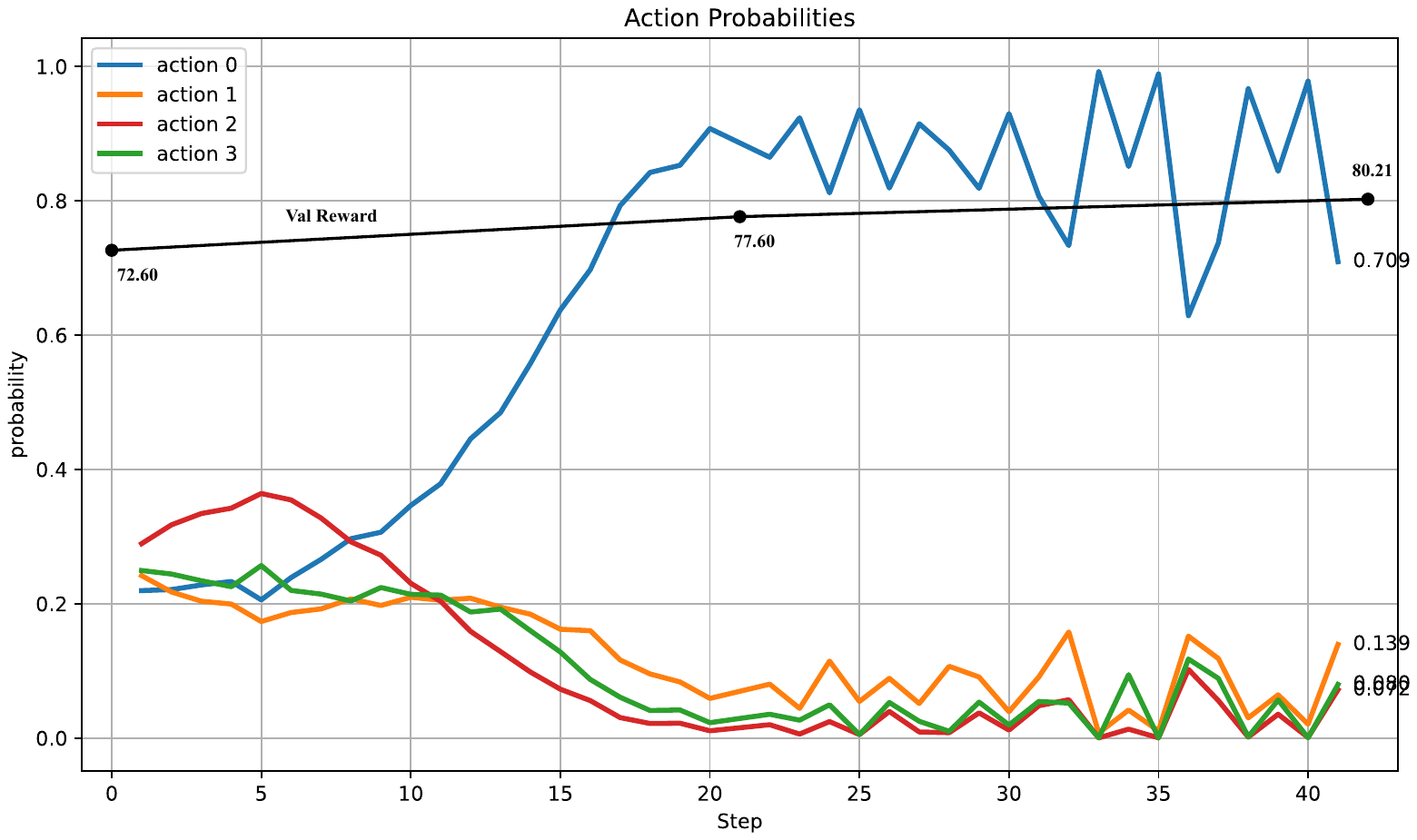}
    \caption{Overall action distribution}
    \label{fig:tok-all}
\end{subfigure}
\caption{\textbf{Action distributions and val reward on MATH} for sampling strategies with the \textbf{token}-level adapter. Actions are defined in \cref{tab:math}. Axes match \cref{fig:math-action-distribution}. Note that a CoT action probability of 0 indicates that no CoT prompt was sampled in that batch.}
\label{fig:token-level-action-distribution}
\end{figure*}

\subsection{Prompts for Qwen3 Models}
\paragraph{MATH.}
\begin{itemize}
    \item \textbf{Without thinking:}  ``Provide the final answer within \textbackslash\textbackslash boxed\{\}. /nothink"
    \item \textbf{With thinking:}  ``Provide the final answer within \textbackslash\textbackslash boxed\{\}. /think"
\end{itemize}

\paragraph{CodeContests.}
\begin{itemize}
    \item \textbf{Without thinking:} ``You are a coding assistant. /nothink"
    \item \textbf{With thinking:} ``You are a coding assistant. Please reason step by step. /think"
\end{itemize}



\end{document}